\documentclass[12pt,draftcls,onecolumn]{IEEEtran} 
\usepackage{multirow}
\usepackage{amsfonts}
\usepackage{amssymb}
\usepackage{amsbsy} 
\usepackage{amsmath}
\usepackage[ansinew]{inputenc}
\usepackage{cite}
\usepackage{pstricks}
\usepackage{multirow}
\usepackage{color}
\usepackage{lscape}
\usepackage[dvips]{graphicx}
\usepackage{url}
\usepackage{epsfig}
\usepackage{subfigure}
\usepackage{algorithmic}
\usepackage{algorithm}
\usepackage{hyperref}

\newcommand{\figdir}{./images/}

\renewcommand{\red}[1]{\textcolor [rgb]{0,0,0}{#1}}
\renewcommand{\blue}[1]{\textcolor [rgb]{0,0,0}{#1}}
\newcommand{\mitch}[2]{\textcolor [rgb]{0,0.0,0}{#2}}
\newcommand{\x}{\mathbf{x}}

\begin{document}

\title{A survey of active learning algorithms for supervised remote sensing image classification}
\author{Devis Tuia,~\IEEEmembership{Member,~IEEE}, 
	Michele Volpi,~\IEEEmembership{Student Member,~IEEE},
	Loris Copa, \\
	Mikhail Kanevski,
	Jordi Mu{\~n}oz-Mar\'i
	\thanks{Manuscript received April 2010;}
	\thanks{This work has been supported by the Swiss National Science Foundation (grants no. 200021-126505 and PBLAP2-127713/1), and by the Spanish 
Ministry of Education and Science under projects TEC2009-
13696, AYA2008-05965-C04-03, and CONSOLIDER/CSD2007-00018.}
	\thanks{DT and JMM are with the Image Processing Laboratory, University of Val\`encia, Val\`encia, Spain. C/ Cat. A. Escardino. 46980 Paterna, Val{\`{e}}ncia, Spain. Email: \{devis.tuia,jordi\}@uv.es, http://ipl.uv.es, Phone: +34 963544021, Fax: +34 963544353}
\thanks{MV and MK are with the Institute of Geomatics and Analysis of Risk, University of Lausanne, Lausanne, Switzerland. Email: \{michele.volpi,mikhail.kanevski\}@unil.ch,  http://www.unil.ch/igar, Phone: +4121-6923546, Fax: +4121-6923535}
\thanks{LC was with the  Institute of Geomatics and Analysis of Risk, University of Lausanne, Lausanne, Switzerland. He is now with SARMAP SA, Switzerland. Email: loris.copa@sarmap.ch, Phone +41 916009365}
	}

\markboth{IEEE Journal of Selected Topics in Signal Processing, preprint. Published version (2011): 10.1109/JSTSP.2011.2139193}
    {Tuia \MakeLowercase{\textit{et al.}}: TITOLO}

\maketitle


\begin{abstract}
\textbf{This is the pre-acceptance version, to read the final version published in 2011 in the IEEE Journal of Selected Topics in Signal Processing (IEEE JSTSP), please go to: \href{https://doi.org/10.1109/JSTSP.2011.2139193}{10.1109/JSTSP.2011.2139193}}\\
Defining an efficient training set is one of the most delicate phases for the success of remote sensing image classification routines. The complexity of the problem, the limited temporal and financial resources, as well as the high intraclass variance can make an algorithm fail if it is trained with a suboptimal dataset. Active learning aims at building efficient training sets by iteratively improving the model performance through sampling.  A user-defined heuristic ranks the unlabeled pixels according to a function of the uncertainty of their class membership and then the user is asked to provide labels for the most uncertain pixels. This paper reviews and tests the main families of active learning algorithms: committee, large margin and posterior probability-based. For each of them, the most recent advances in the remote sensing community are discussed and some heuristics are detailed and tested. Several challenging remote sensing scenarios are considered, including very high spatial resolution and hyperspectral image classification. Finally, guidelines for choosing the good architecture are provided for new and/or unexperienced user.
\end{abstract}


\section{Introduction}

\PARstart{N}{owadays}, the recourse to statistical learning models~\cite{Has09} is  a common practice for remote sensing data users; models such as Support Vector Machines (SVM,~\cite{Bos92,Sch02}) or neural networks~\cite{Hay08} are considered as state of the art algorithms for the classification of landuse using  new generation satellite imagery~\cite{Cam09}. Applications of such models to very high spatial~\cite{bru06b,Pac08c,Tui09a} or spectral~\cite{Mel04,Cam04,Fau08b} resolution have proven their efficiency for handling remote sensing data.

However, the performances of supervised algorithms strongly depend on the representativeness of the data used to train the classifier~\cite{Foo04}. This constraint makes the generation of an appropriate training set a difficult and expensive task requiring extensive manual analysis of the image. This is usually done by visual inspection of the scene or by field surveys and successive labeling of each sample.

In the case of field surveys, which is usual for medium resolution, hyperspectral or SAR images, the discovery of a new label is expensive -- both in terms of time and money -- because it involves terrain campaigns. Therefore, there is a limit to the number of pixels that can be acquired. For this reason, compact and informative training sets are needed. 

In the case of visual inspection or photo-interpretation, more common in VHR imagery, it is easier to collect data samples, since the labeling can be done directly on the image. However, the labeling is often done by mass selection on screen and several neighboring pixels carrying the same information are included. As a consequence, the training set is highly redundant. Such a redundancy considerably slows down the training phase of the model. \mitch{, because several pixels carrying the same information are evaluated.}{} Moreover,  the inclusion of noisy pixels may result in a wrong representation of the class statistics, which may lead to poor classification performances and/or overfitting~\cite{Foo06a}. For these reasons, a training set built by photointerpretation should also be kept as small as possible and focused on those pixels effectively improving the performance of the model.

Summing up, besides being small, a desirable training set must be constructed in a smart way, meaning it must represent correctly the class \mitch{statistics and covering the entirety of the data variability}{boundaries by sampling discriminative pixels}. This is particularly critical in very high spatial and spectral resolution image classification, which deal with large \mitch{}{and/or} complex features spaces using limited training information only~\cite{Foo06}. 

In the machine learning literature this approach to sampling is known as \emph{active learning}. The leading idea is that a classifier trained on a small set of well-chosen examples can perform  as well as a classifier trained on a larger number of randomly chosen examples, while it is computationally cheaper~\cite{Mac92,Coh94,Coh96}. Active learning focuses on the interaction between the user and the classifier. In other words, the model returns to the user the  pixels whose classification outcome is the most uncertain. After accurate labeling by the user, pixels are included into the training set in order to reinforce the model. This way, the model is optimized on well-chosen difficult examples, maximizing its generalization capabilities.

The active learning framework has demonstrated its effectiveness when applied to large datasets needing accurate selection of examples~\cite{Lew94}. This is suitable for remote sensing applications, where the number of pixels among which the search is performed is large and manual definition is - as stated above - redundant and time consuming.
As a consequence, active learning algorithms gain an increasing interest in remote sensing image processing and several approaches have been proposed to solve image classification tasks. This paper presents the general framework of active learning and reviews some of the methods that have been proposed in remote sensing literature. Note that this survey only covers remote sensing application of active learning principles: for a general introduction and survey of the most recent developments in the machine learning community, please refer to~\cite{Ols09,Set10}.

The remainder of the paper is organized as follows: Section~\ref{sec:AL} presents the general framework of active learning and the families of methods that will be detailed in \red{Sections~\ref{sec:committee} to~\ref{sec:poster}}, as well as the references to specific methods. Section~\ref{sec:data} presents the datasets considered in the experiments. Section~\ref{sec:res} compares the different approaches numerically. Section~\ref{sec:disc} gives an overview and guidelines for potential users. Section~\ref{sec:concl} concludes the paper.

\section{Active learning: concepts and definitions}\label{sec:AL}
Let $X~=~\{\x_i,y_i\}_{i=1}^l$ be a training set of labeled samples, with $\x_i \in \mathcal{X}$ and $y_i = \{1, ...,  N\}$. $\mathcal{X}$ is the $d$-dimensional input space $\in \mathbb{R}^d$. Let also $U~=~\{\x_i\}_{i=l+1}^{l+u} \in \mathcal{X}$, with $u \gg l$ be the set of unlabeled pixels to be sampled, or the \emph{pool of candidates}. 

Active learning algorithms are iterative sampling schemes, where a classification model is adapted regularly by feeding it with new labeled pixels corresponding to the ones that are most beneficial for the improvement of the model performance. These pixels are usually found in the areas of \emph{uncertainty} of the model and their inclusion in the training set forces the model to solve the \mitch{uncertainties}{regions of low confidence}. For a given iteration $\epsilon$, the algorithm selects from the pool $U^\epsilon$ the $q$ candidates that will at the same time maximize the gain in performance and reduce the uncertainty of the model when added to the current training set $X^\epsilon$. Once the batch of pixels $S^\epsilon = \{\x_m\}_{m=1}^q \subset U$ has been selected, it is labeled by the user, i.e. the labels $\{y_m\}_{m=1}^q$ are discovered. Finally, the set $S^\epsilon$ is both added to the current training set ($X^{\epsilon+1} = X^\epsilon \cup S^\epsilon$)  and removed from the pool ($U^{\epsilon+1} = U^\epsilon \backslash S^\epsilon$). The process is iterated until a stopping criterion is met. Algorithm~\ref{alg:AL} summarizes the active selection process. From now on, the iteration index $\epsilon$ will be omitted in order to ease notation.

\begin{algorithm}[!t]
\caption{General active learning algorithm}
\label{alg:AL}
\vspace{0.2cm}
\textbf{Inputs}\\
- Initial training set $X^\epsilon = \{\x_i,y_i\}_{i=1}^l$ ($X \in \mathcal{X}$, $\epsilon=1$).\\
- Pool of candidates $U^\epsilon = \{\x_i\}_{i=l+1}^{l+u}$ ($U \in \mathcal{X}$, $\epsilon=1$).\\
- Number of pixels $q$ to add at each iteration (defining the batch of selected pixels $S$).
\vspace{0.2cm}
\begin{algorithmic}[1]
\REPEAT
	\STATE Train a model with current training set $X^\epsilon$.
	\FOR{each candidate in $U^\epsilon$}
	\STATE Evaluate a user-defined \emph{heuristic}
	\ENDFOR
	\STATE Rank the candidates in $U^\epsilon$ according to the score of the heuristic
	\STATE Select the $q$ most interesting pixels.  $S^\epsilon = \{\x_k\}_{k=1}^q$
	\STATE The user assigns a label to the selected pixels.  $S^\epsilon = \{\x_k,y_k\}_{k=1}^q$
	\STATE Add the batch to the training set $X^{\epsilon+1}=X^{\epsilon} \cup S^\epsilon$.
	\STATE Remove the batch from the pool of candidates $U^{\epsilon+1}=U^{\epsilon} \backslash S^\epsilon$
	\STATE $\epsilon = \epsilon + 1$
			
\UNTIL{a stopping criterion is met.}
\end{algorithmic}
\vspace{0.2cm}
\end{algorithm}

An active learning process requires interaction between the user and the model: the first provides the labeled information and the knowledge about the desired \blue{classes}, while the latter provides both its own interpretation of the distribution of the classes and the most relevant pixels that are needed  in order to solve the discrepancies encountered. This point is crucial for the success of an active learning algorithm: the machine needs a strategy to rank the pixels in the pool $U$. 
These strategies, or \emph{heuristics}, differentiate the algorithms proposed in the next sections and can be divided into three main families~\cite{Tui09}: 

\begin{itemize}
	\item[1 -] \emph{Committee}-based heuristics (Section~\ref{sec:committee})
	\item[2 -] \emph{Large margin}-based heuristics (Section~\ref{sec:lm})
	\item[3 -] \emph{Posterior probability}-based heuristics (Section~\ref{sec:poster})
\end{itemize}

A last family of active learning heuristics, the \emph{cluster}-based, has recently being proposed in the community~\cite{Tui10}: cluster-based heuristics aim at pruning a hierarchical clustering tree until the resulting clusters are consistent with the labels provided by the user. Therefore, these strategies rely on an unsupervised model, rather than on a predictive model. Since the aim of these heuristics is different form that of the other families presented, they will not be detailed in this survey.

\section{Committee based active learning}
\label{sec:committee}
The first family of active learning methods quantifies the uncertainty of a pixel  by considering a committee of learners~\cite{Seu92,Fre97}. Each member of the committee exploits different hypotheses about the classification problem and consequently labels the pixels in the pool of candidates. The algorithm then selects the samples showing maximal disagreement between the different classification models in the committee.  
To limit computational complexity, heuristics based on multiple classifier systems have been proposed in machine learning literature. In~\cite{Abe98}, methods based on boosting and bagging are proposed in this sense for binary classification only. In~\cite{Mel04c}, results obtained by query-by-boosting and query-by-bagging are compared using several batch datasets showing excellent performance of the heuristics proposed. Methods of this family have the advantage to be applicable to any kind of model or combination of  models. 
In the remote sensing community, committee-based approaches to active learning have been proposed exploiting two types of uncertainty: first, committees varying the pixels members have been considered in the query-by-bagging heuristic~\cite{Tui09,Cop10}. Then, committees based on subsets of the feature space available have been presented in~Di and Crawford~\cite{Di10b}. The next two sections present the algorithms proposed in these papers.

\subsection{Normalized entropy query-by-bagging (nEQB)}\label{sec:EQB}
In the implementations of \cite{Abe98}, bagging  \cite{Bre94} is proposed to build the committee: first, $k$ training sets built on a draw with replacement of the original data are defined. These draws account for a part of the available labeled pixels only. Then, each set is used to train a classifier and to predict the $u$ labels of the candidates. At the end of the procedure, $k$ labelings are provided for each candidate pixel $\x_i \in U$.
In~\cite{Tui09}, the entropy $H^{\text{BAG}}$ of the distribution of the predictions \mitch{provided by the $k$ classifiers for each pixel $\x_i$ in $U$}{} is used as heuristic. In~\cite{Cop10}, this measure has been subsequently normalized in order to bound it with respect to the number of classes predicted by the committee and avoid hot spots of the value of uncertainty in regions where several classes overlap. The \emph{normalized entropy query-by-bagging} heuristic can be stated as follows:

\begin{equation}
	\hat{\x}^{\text{\emph{n}EQB}} = \arg\max_{\x_i \in U}\Big\{\frac{H^{\text{BAG}}(\x_i)}{ \mbox{log}(N_i)}\Big\}
\label{eq:neqb}
\end{equation}

where

\begin{align}
	&H^{\text{BAG}}(\x_i) = -\sum_{\omega=1}^{N_i} p^{\text{BAG}}(y_i^* = \omega|\x_i) \mbox{log}\left[p^{\text{BAG}}(y^*_i= \omega|\x_i)\right]\\
		& \text{where} \qquad p^{\text{BAG}}(y^*_i = \omega|\x_i) = \frac{\sum_{m=1}^k \delta(y_{i,m}^{*}, \omega )}{\sum_{m=1}^k\sum_{j = 1}^{N_i} \delta(y_{i,m}^*, \omega_j ) } \nonumber 
\label{eq:entropy}
\end{align}

$H^{\text{BAG}}(\x_i)$ is an empirical measure of entropy, $y_i^*$ is the prediction for the pixel $\x_i$ and $p^{\text{BAG}}(y^*_i = \omega|\x_i)$ is the observed probability to have the class $\omega$ predicted using the training set $X$ by the committee of $k$ models for the sample $\x_i$. $N_i$ is the number of classes predicted for pixel $\x_i$ \mitch{by the committee}{}, with $1 \leq N_i \leq N$. The $\delta(y_{i,m}^*,\omega)$ operator returns the value $1$ if the classifier using the $m$-th bag classifies the sample $\x_i$ into class $\omega$ and $0$ otherwise. Entropy maximization gives a naturally multiclass heuristic. A candidate for which all the classifiers in the committee agree is associated with null entropy and its inclusion in the  training set does not bring additional information. On the contrary, a candidate with maximum disagreement between the classifiers results in maximum entropy, and its inclusion will be highly beneficial.

\subsection{Adaptive maximum disagreement (AMD)}
When confronted to high dimensional data, it may be relevant to construct the committee by splitting the feature space into a number of subsets, or \emph{views}~\cite{Mus06}. Di and Crawford~\cite{Di10b} exploit this principle to generate different views of a hyperspectral image on the basis of the block-diagonal structure of the covariance matrix. By generating views corresponding to the different blocks, independent classifications of the same pixel can be generated and an entropy-based heuristic can be used similarly to $n$EQB.

Given a partition of the $d$-dimensional input space into $V$ disjoint views accounting for data subsets $\x^v$ such that $\bigcup_{v=1}^V \x^v = \x $, the `Adaptive maximum disagreement ' (AMD) heuristic selects candidates according to the following rule:

\begin{equation}
	\hat{\x}^{\text{AMD}} = \arg\max_{\x_i \in U} H^{\text{MV}}(\x_i)
	\label{eq:AMD}
\end{equation}

where the multiview entropy $H^{\text{MV}}$ is assessed over the predictions of classifiers using a specific view $v$:

\begin{align}
	&H^{\text{MV}}(\x_i) =  -\sum_{\omega=1}^{N_i} p^{\text{MV}}(y^*_{i,v} = \omega|\x^v_i) \mbox{log}\left[p^{\text{MV}}(y^*_{i,v}= \omega|\x^v_i)\right]\\
	& \text{where} \qquad p^{\text{MV}}(y^*_i = \omega|\x^v_i) = \frac{\sum_{v=1}^V W^{\epsilon-1}(v,\omega) \delta(y^*_{i,v}, \omega )}{\sum_{v=1}^V\sum_{j = 1}^{N_i} W^{\epsilon-1}(v,\omega)} \nonumber
\label{eq:MVentr}
\end{align}

where the $\delta(y_{i,v}^*,\omega )$ operator returns the value $1$ if the classifier using the view $v$ classifies the sample $y_i$ into class $\omega$ and $0$ otherwise. $\mathbf{W}^{\epsilon-1}$ is a $N \times V$ weighting matrix incorporating the abilities of discrimination between the views in the different classes. At each iteration, $\mathbf{W}^{\epsilon-1}$ is updated on the basis of the true labels of the pixels sampled at iteration $\epsilon-1$:

\begin{equation}
W^\epsilon(v,\omega) = W^{\epsilon-1}(v,\omega)+\delta(y_{i,v},\omega ), \quad \forall i \in S
\label{eq:W}
\end{equation}

and its columns are normalized to a unitary sum. This matrix weights the confidence of each view to predict a given class. In~\cite{Di10b}, the selection is done on a  subset of $U$ containing the candidate pixels maximizing the uncertainty, which are the pixels for which  the committee has predicted the highest number of classes. This way, the computational load of the algorithm is reduced.

\section{Large margin based active learning}
\label{sec:lm}
The second family of methods is specific to margin-based classifiers. 
Methods such as SVM are naturally good base methods for active learning: the distance to the separating hyperplane, that is the absolute value of the decision function without the sign operator, is a straightforward way of estimating the classifier confidence on an unseen sample. Let's first consider a binary problem: the distance of a sample $\x_i$ from the SVM hyperplane is given by

\begin{equation}
f(\x_i)=\sum_{j=1}^n \alpha_j y_j K(\x_j,\x_i)+b
\label{eq:decfunct}
\end{equation}

where $K(\x_j, \x_i)$ is a kernel, which defines the similarity between the candidate $\x_i$ and the support vectors $\x_j$, which are the pixels showing non zero $\alpha_j$ coefficients. The labels $y_j$ of the support vectors are $+1$ for samples of the positive class and $-1$ for those on the negative. For additional information, see SVM literature in~\cite{Bos92,Sch02}. 

This evaluation of the distance is the base ingredient of almost all large margin heuristics. Roughly speaking, these heuristics use the intuition that a sample away from the decision boundary (with a high $f(\x_i)$) has a high confidence about its class assignment and is thus not interesting for future sampling.

Since SVM rely on a sparse representation of the data, large margin based heuristics aim at finding the pixels in $U$ that are most likely to receive a non-zero $\alpha_i$ weight if added to $X$. In other words, the points more likely to become support vectors are the ones lying within the margin of the current model~\cite{Ton01a}. The heuristic taking advantage of this property is called margin sampling (MS)~\cite{Cam00,Sch00}. Recent modifications of MS, aiming at minimizing the risk of selecting points that will not become support vectors, can be found in~\cite{Zom04,Che07a}.
MS is the most studied active learning algorithm in remote sensing. Its first application can be found in~\cite{Mit04}. Modifications of the MS heuristic have been proposed in~\cite{Bru09c}. Later on, since no cross-information among the samples is considered in the MS, the questions of diversity in batches of samples have been considered in~\cite{Fer07,Tui09,Dem10}. The next sections present the MS heuristic and subsequent modifications proposed in order to enhance diversity when selecting batches of samples.

\subsection{Margin sampling (MS)}
\begin{figure*}
\centering
\begin{tabular}{|ccc|}
\hline
\includegraphics[width=4.2cm]{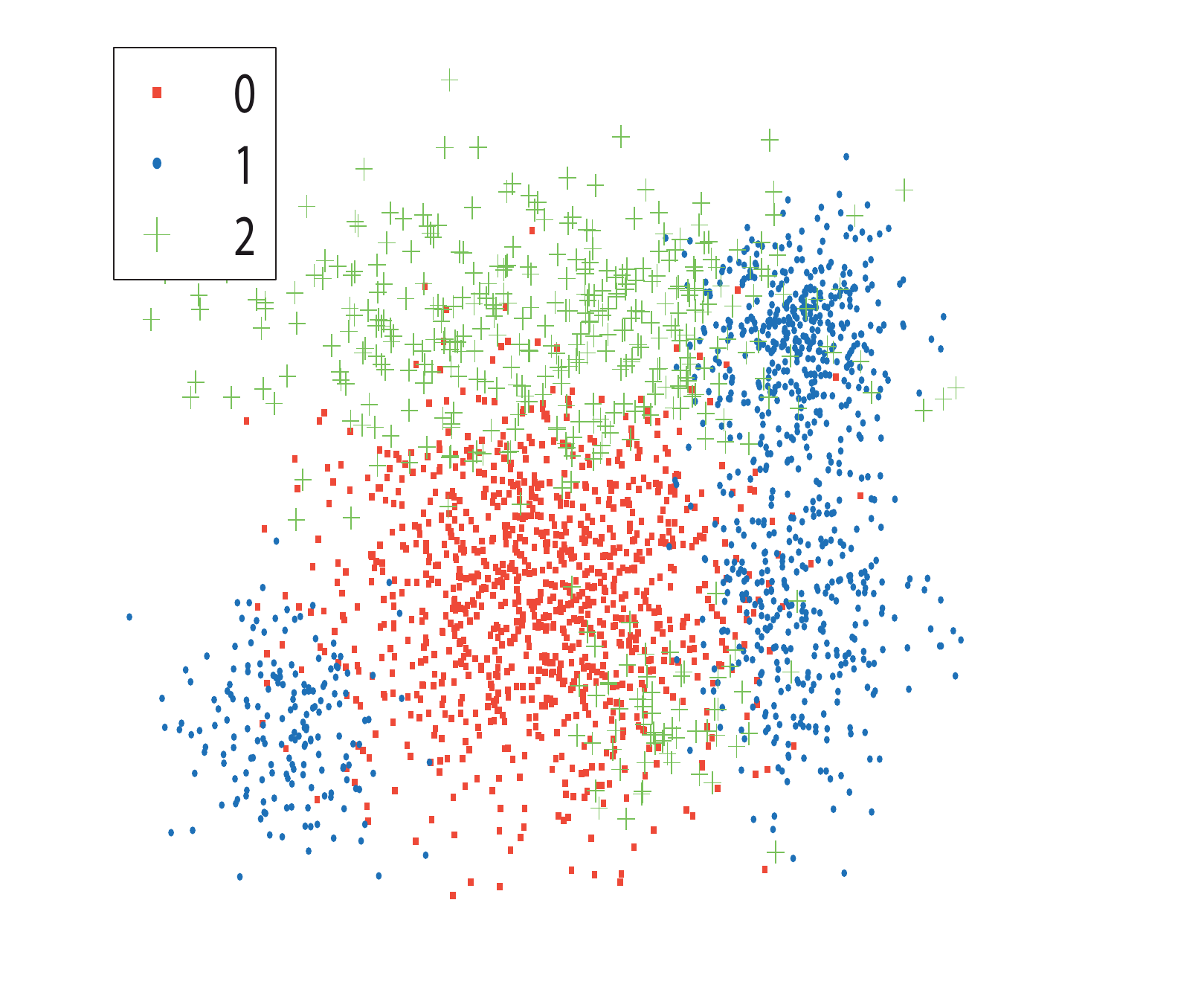}&
\includegraphics[width=4.2cm]{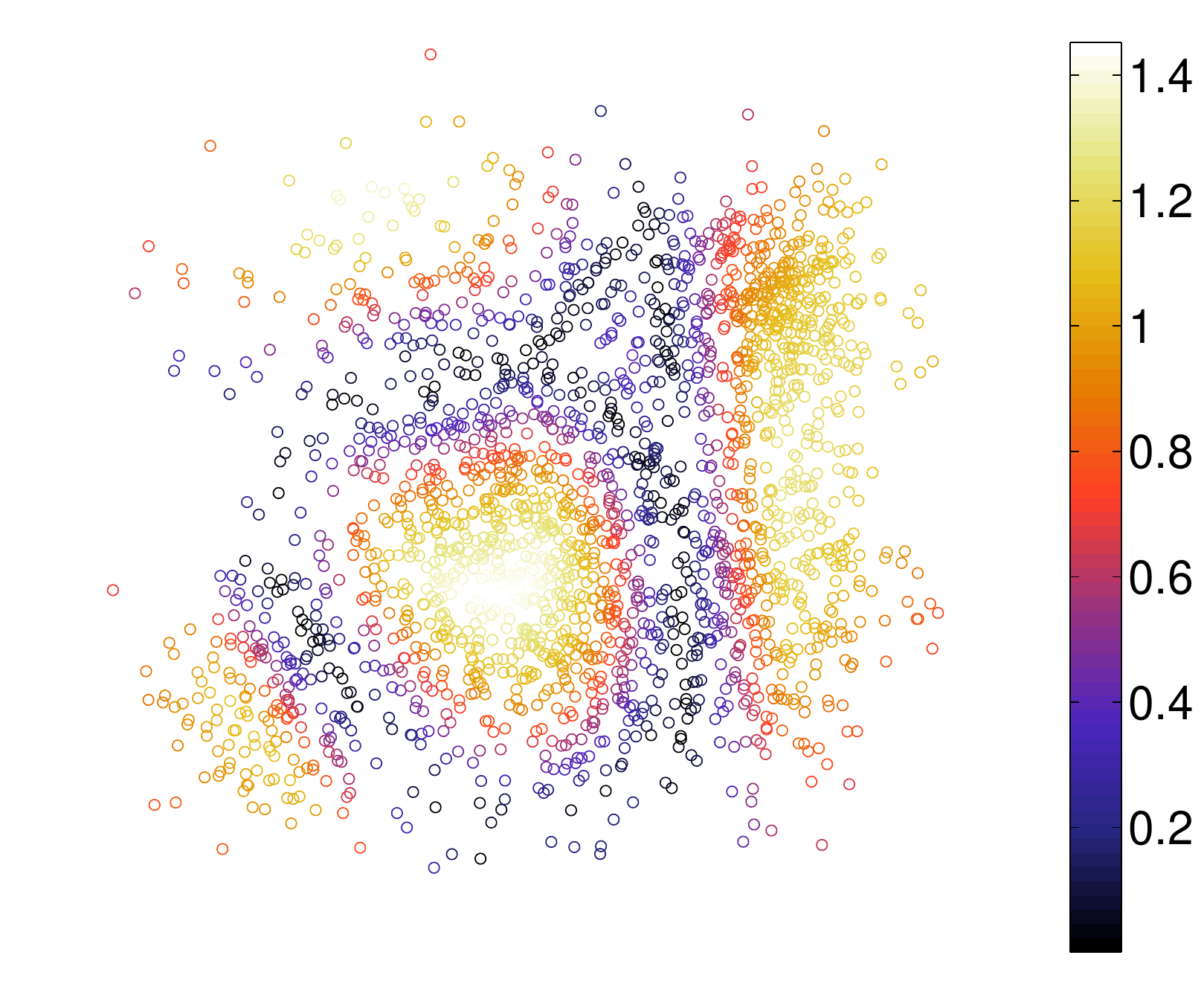}&
\includegraphics[width=4.2cm]{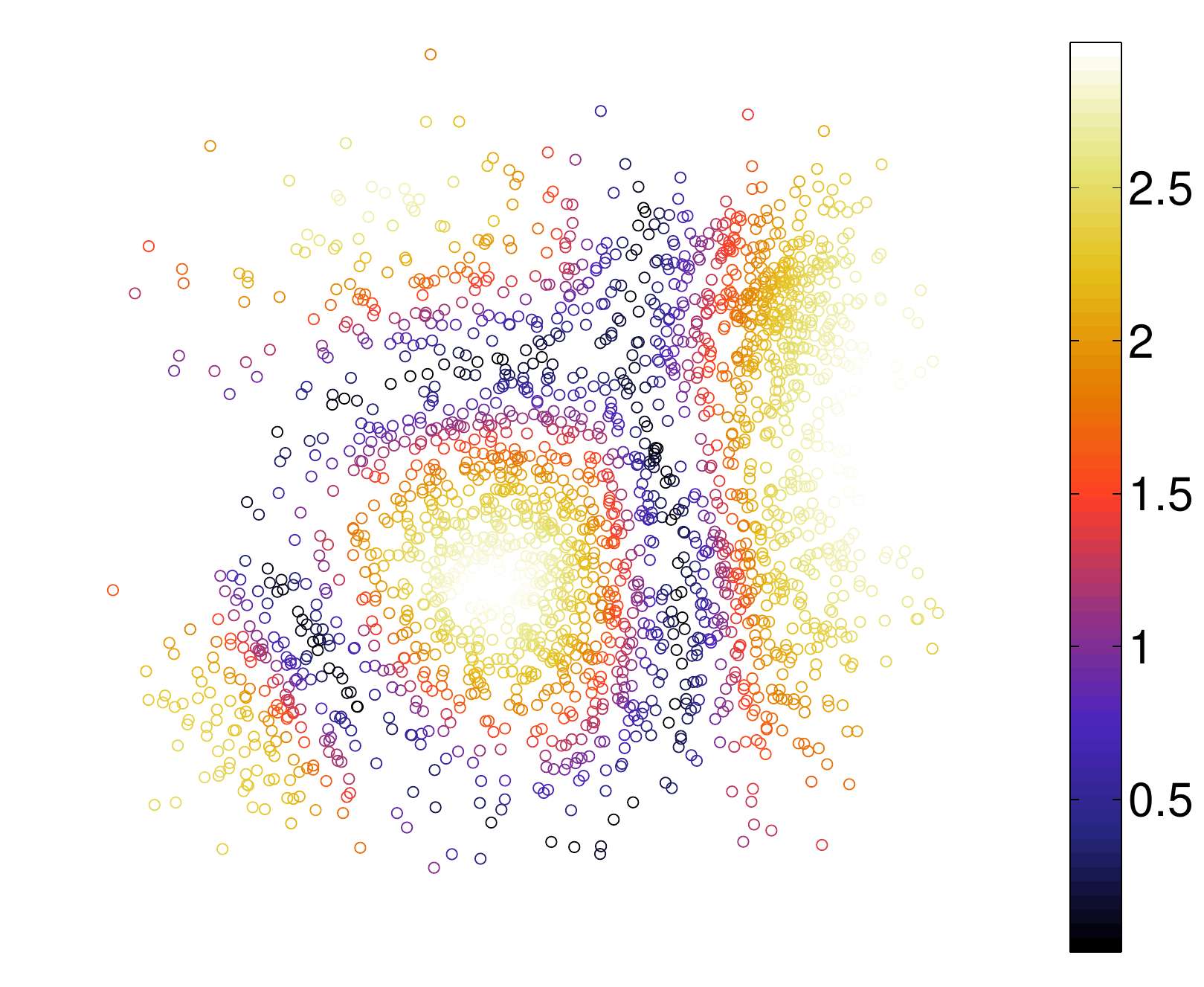}\\
(a)&(b) MS, Eq.~\eqref{eq:minMarg}&(c) MCLU, Eq.~\eqref{eq:MCLUfct}\\
\includegraphics[width=4.2cm]{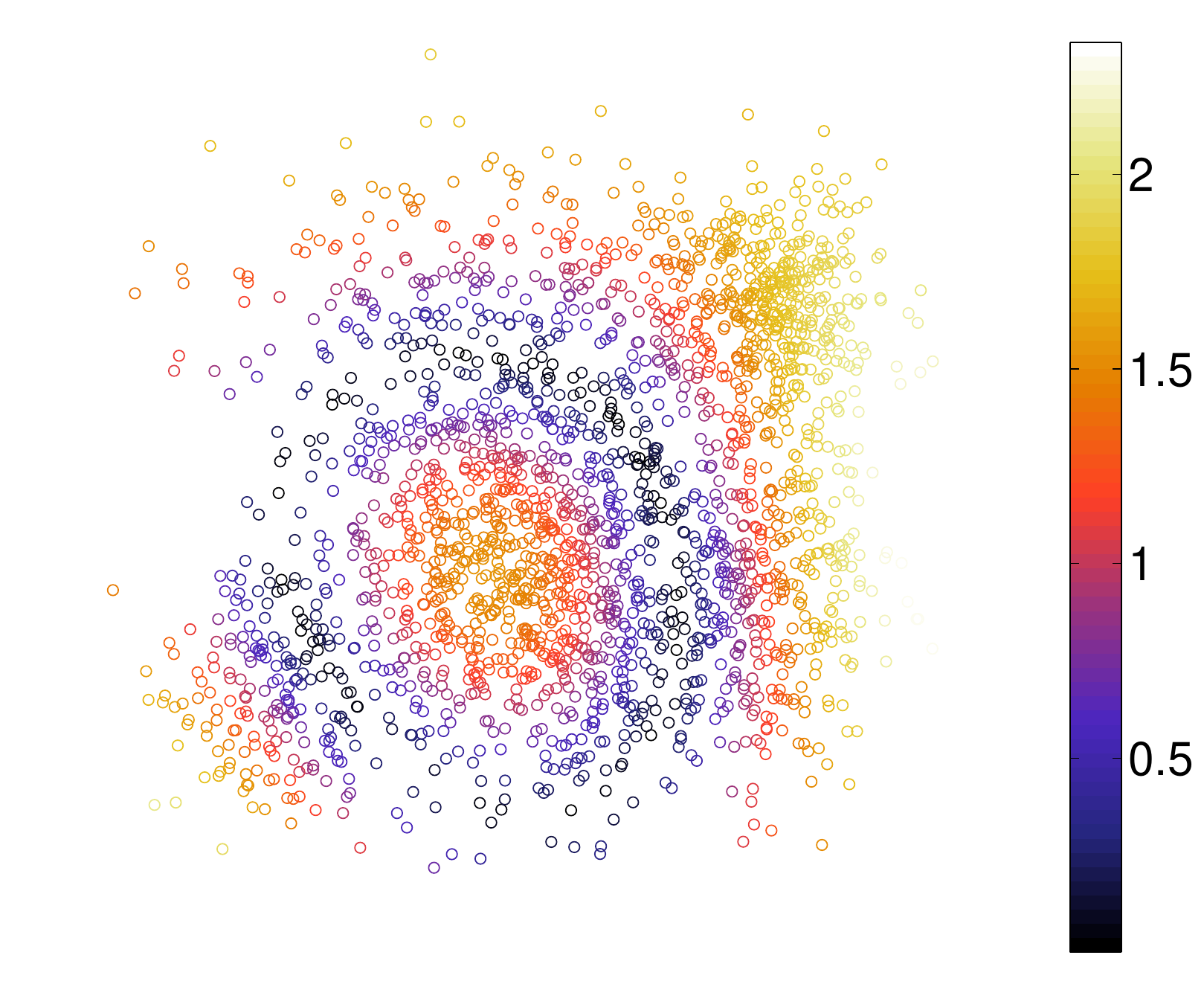}&
\includegraphics[width=4.2cm]{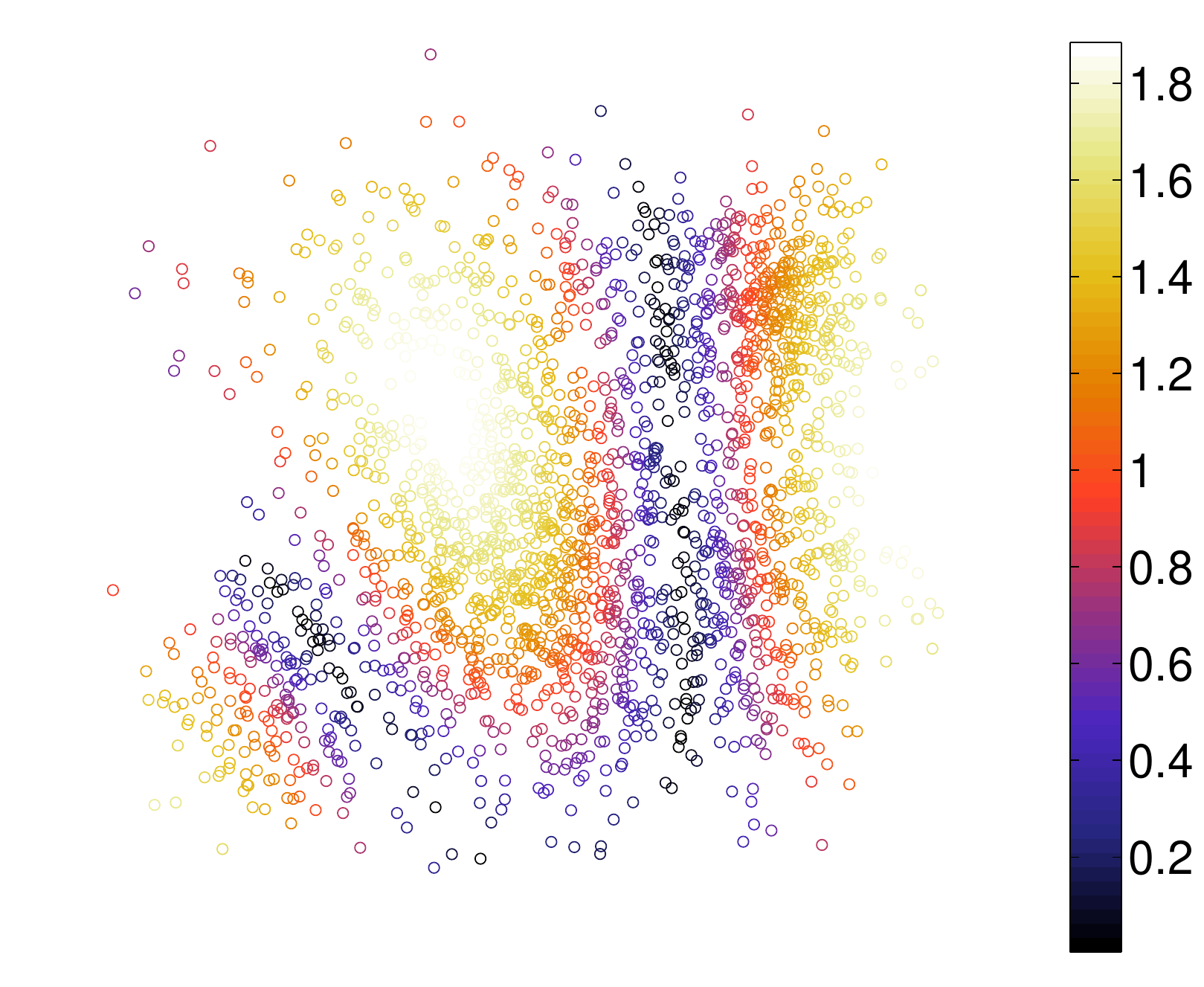}&
\includegraphics[width=4.2cm]{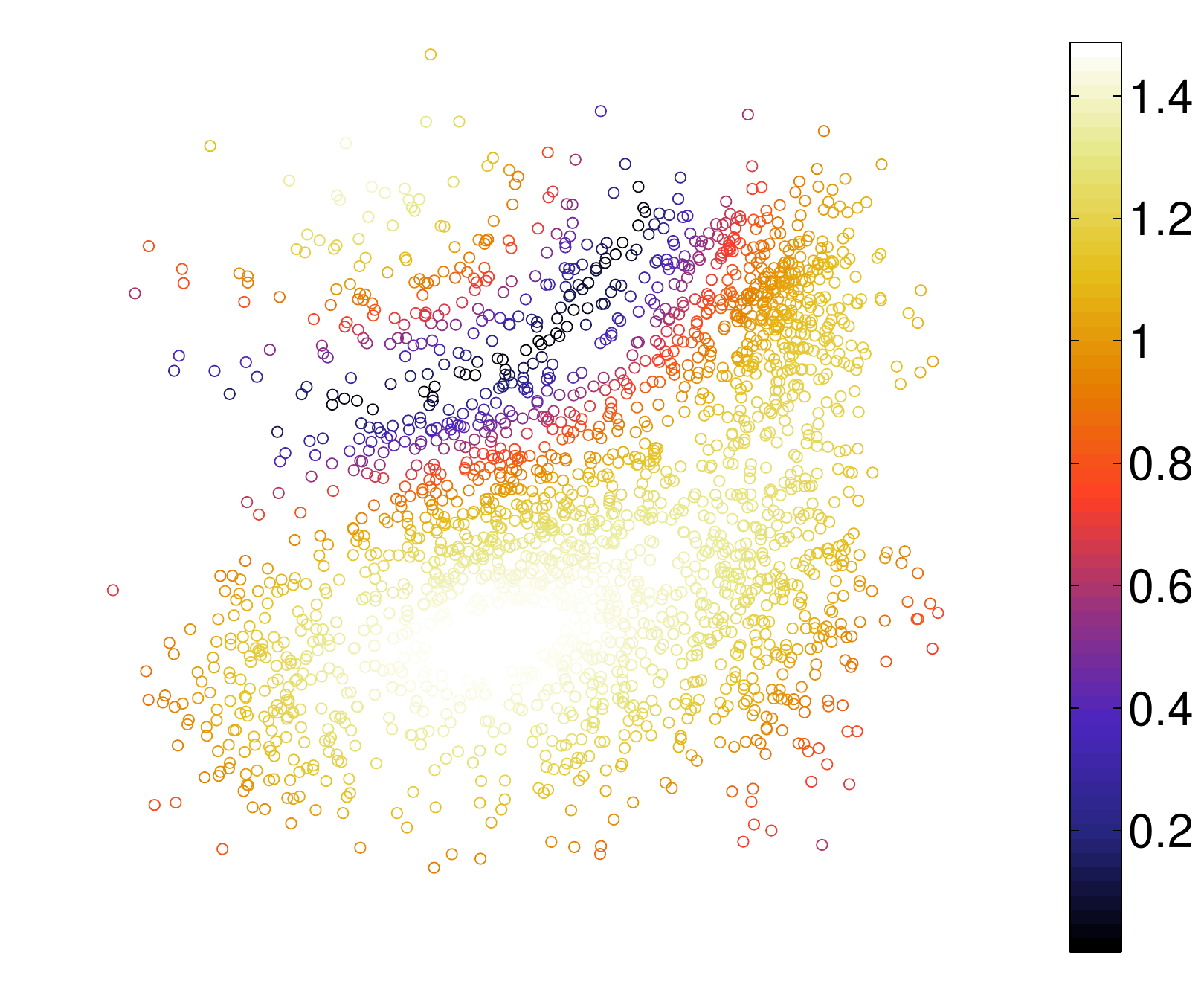}\\
(d) $|f(\x,\omega_0)|$&(e) $|f(\x,\omega_1)|$&(f) $|f(\x,\omega_2)|$\\ \hline
\end{tabular}
\caption{Large margin heuristics for a three classes toy example represented in subfigure (a). The color intensity represents the distance from the hyperplane, ranging from black (on the boundary) to white (maximal distance): (b) MS heuristic; (c) MCLU heuristic; areas in black are the areas of maximal uncertainty, minimizing Eq.~\eqref{eq:minMarg} or Eq.~\eqref{eq:MCLUfct} respectively. 
Bottom row:  absolute values of per-class distances (d)-(f). }
\label{fig:dist}
\end{figure*}

As stated above, margin sampling  takes advantage of SVM geometrical properties, and in particular of the fact that unbounded support vectors are labeled examples that lie on the margin with a decision function value of exactly one~\cite{Bos92,Sch02}. Consider the pool of candidates of Fig.~\ref{fig:dist}(a) referring to a three classes toy problem. In a multiclass one-against-all setting, the distance to each hyperplane is represented by Figs.~\ref{fig:dist}(d-f). 
The `margin sampling' (MS) heuristic performs sampling of the candidates by minimizing Eq.~\eqref{eq:minMarg}:

\begin{equation}
\hat{\x}^{\text{MS}} = \arg\min_{\x_i \in U } \Big\{\min_{\omega}|f(\x_i,\omega)|\Big\}
\label{eq:minMarg}
\end{equation}

where $f(\x_i,\omega)$ is the distance of the sample to the hyperplane defined for class $\omega$ in a one-against-all setting for multiclass problems. The MS heuristic for the toy problem is reported in~Fig.~\ref{fig:dist}(b). MS heuristic can be found in the literature under the names of `most ambiguous'~\cite{Fer07}, `binary level uncertainty'~\cite{Dem10} or SVM$_\text{SIMPLE}$~\cite{Di10b}.

\subsection{Multiclass level uncertainty (MCLU)}\label{sec:MCLU}
In~\cite{Dem10}, the idea of margin sampling is extended to multiclass uncertainty (see~\cite{Vla08}). Instead of dealing with the most uncertain class of the SVM, the `multiclass level uncertainty' (MCLU) considers the difference between the distance to the margin for the two most probable classes. 

\begin{align}
&\hat{\x}^{\text{MCLU}} = \arg \min_{\x_i \in U} \Big\{f(\x_i)^{\text{MC}}\Big\}\label{eq:MCLU}\\
&\text{where} \qquad f(\x_i)^{\text{MC}} =  \max_{\omega \in N}|f(\x_i,\omega)|-\max_{\omega \in N\backslash \omega^+}|f(\x_i,\omega)|\label{eq:MCLUfct}
\end{align}

where $\omega^+$ is the class showing maximal confidence, i.e. the argument of the first term of Eq.~\eqref{eq:MCLUfct} showing maximal $f(\x_i)^{\text{MC}}$. A high value of this criterion corresponds to samples assigned with high certainty to the most confident class, while a small value represents unreliable classification. Fig.~\ref{fig:dist}(c) illustrates the heuristic in comparison to MS. Although they are very similar, MCLU performs better in the area where the three classes mix, in the top-right area of the feature space: in this area, MCLU returns maximal uncertainty as it is evaluated on the basis of all the per-class decision values, while MS returns an uncertainty slightly lower than on the two-classes boundaries.

\subsection{Significance space construction (SSC)}
In~\cite{Pas10b}, instead of using the distance to the hyperplane as a measure of uncertainty, the support vector coefficients are used to convert the multiclass classification problem into a binary support vector detection problem. In the `significance space construction' (SSC) heuristic, the training samples related to support vector coefficients are used to define a second classification function $f(\x)^{\text{SSC}}$, where training pixels with $\alpha_j > 0$ (the support vectors) are classified against training pixels with $\alpha_j = 0$. Once applied to the pool of candidates, this second classifier predicts which pixels are likely to become support vectors. 

\begin{equation}
\hat{\x}^{\text{SSC}} = \arg_{\x_i \in U}  f(\x_i)^{\text{SSC}} > 0
\label{eq:SSC}
\end{equation}

Once the candidates more likely to become support vectors have been highlighted, a random selection among them is done.

\subsection{On the need for a diversity criterion}
In applicative scenarios, diversity among samples~\cite{Bri03} is highly desirable. Diversity concerns the capability of the model to reject candidates that rank well according to the heuristic, but are redundant among each other. Diversity has been studied extensively for margin-based heuristics, where the base margin sampling heuristic is constrained using a measure of diversity between the candidates (see Algorithm~\ref{alg:diversity}). 

\begin{algorithm}[!b]
\caption{General diversity based heuristic (for a single iteration)}
\label{alg:diversity}
\vspace{0.2cm}
\textbf{Inputs}\\
- Current training set $X^{\epsilon} = \{\x_i,y_i\}_{i=1}^l$ ($X \in \mathcal{X}$).\\
- Subset of the pool of candidates minimizing Eq.~\eqref{eq:minMarg} or~\eqref{eq:MCLU} $U^* = \{\x_i\}$ $ (U^* \in \mathcal{X}$ and $U^* \subset U^{\epsilon}$).\\
- Number of pixels $q$ to add at each iteration (defining the batch of selected pixels $S$).
\vspace{0.2cm}
\begin{algorithmic}[1]

\STATE Add the pixel minimizing Eq.~\eqref{eq:minMarg} or~\eqref{eq:MCLUfct} to $S$.

\REPEAT{}
	\STATE Compute the user defined diversity criterion between pixels in $U^*$ and in $S$ (with MAO, cSV or ABD).
	\STATE Select the pixel $\x_D$ maximizing diversity with current batch.
	\STATE Add $\x_D$ to current batch $S = S \cup \x_D$.
	
	\STATE Remove $\x_D$ to current list of cadidates $U^* = U^* \setminus \x_D$.
				
\UNTIL{batch $S$ contains $q$ elements.}
\STATE The user labels the selected pixels.  $S = \{\x_k,y_k\}_{k=1}^q$
\STATE Add the batch to the training set $X^{\epsilon+1}=X^{\epsilon} \cup S$.
\STATE Remove the batch from the complete pool of candidates $U^{\epsilon+1}=U^{\epsilon} \backslash S$

\end{algorithmic}
\vspace{0.2cm}
\end{algorithm}

The first heuristic proposing explicit diversity in remote sensing is found in~\cite{Fer07}, where the margin sampling heuristic is constrained with a measure of the angle between candidates in feature space. This heuristic, called `most ambiguous and orthogonal' (MAO) is iterative: starting from the samples selected by MS, \mitch{}{$U^{\text{MS}}\subset U$}, this heuristic iteratively chooses the samples minimizing the highest values between the candidates list and the samples already included in the batch \mitch{of samples}{} $S$. For a single iteration, this can be resumed as:

\begin{equation}
\hat{\x}^{\text{MAO}} = \arg \min_{\x_i \in U^{\text{MS}} }\Big\{\max_{\x_j \in S}K(\x_i,\x_j)\Big\}
\label{eq:MAO}
\end{equation}

In~\cite{Dem10}, the MAO criterion is combined with the MCLU uncertainty estimation in the `multiclass level uncertainty - angle-based diversity' (MCLU-ABD) heuristic. Te selection is performed among a subset of $U$ maximizing the MCLU criterion. Moreover, the author generalize the MAO heuristic to any type of kernels by including normalization in feature space.

\begin{align}
\hat{\x}^{\text{MCLU-ABD}} = \arg \min_{\x_i \in U^{\text{MCLU}} }&\Bigg\{\lambda f(\x_i)^{\text{MC}} + \\\nonumber
&(1-\lambda)\max_{\x_j \in S}\frac{K(\x_i,\x_j)}{\sqrt{K(\x_i,\x_i)K(\x_j,\x_j)}}\Bigg\}\\
\label{eq:MCLU-ABD}
\end{align}

where $f(\x_i)^{\text{MC}}$ is the multiclass uncertainty function defined by Eq.~\eqref{eq:MCLUfct}.

In~\cite{Tui09}, the diversity of the chosen candidates is enforced by constraining the MS solution to pixels associated to different closest support vectors. This approach ensures a certain degree of diversification in the MS heuristic, by dividing the margin in the feature space as a function of the geometrical distribution of the support vectors. Compared to the previously presented heuristics, this approach has the advantage of ensuring diversity with respect to the current model, but does not guarantee diversity of the samples between each other (since two close samples can be associated to different support vectors).

\begin{equation}
\hat{\x}^{\text{cSV}} = \arg \min_{\x_i \in U^{\text{MS}} } \Big\{\lvert f(\x_i,\omega)\rvert \   \Big\vert  \   cSV_i \not\in cSV_\theta\Big\}
\label{eq:cSV}
\end{equation}

where $\theta = [1, \ldots, q-1]$ are the indices of the already selected candidates and $cSV$ is the set of selected closest support vectors.

Finally, diversity can be ensured using clustering in the feature space. In~\cite{Dem10}, kernel $k$-means~\cite{Gir02, Dhi05, Sha04} was used to cluster the samples selected by MCLU and select diverse batches. After partitioning the $U^{\text{MCLU}}$ set into $q$ clusters with kernel $k$-means, the `multiclass level uncertainty - enhanced cluster based diversity (MCLU-ECBD)' selects a single pixel per cluster, minimizing the following query function:

\begin{equation}
\hat{\x}^{\text{MCLU-ECBD}} = \arg \min_{\x_i \in c_m}\Big\{ f(\x_i)^{\text{MC}}\Big\}\text{,} \quad m = [1, ... q],  \x_i \in U^{\text{MCLU}}
\label{eq:MCLU-ECBD}
\end{equation}

where $c_m$ is one among the $q$ clusters defined with kernel $k$-means. 

In~\cite{Vol10d}, a hierarchical extension of this principle is proposed to exclude from the selected batch the pixels more likely to become bounded support vectors. This way, the redundancy affecting samples close to each other in the feature space among different iterations is controlled along with the maximization of the informativeness of each pixel.
In the `informative hierarchical margin cluster sampling' (hMCS-i), a dataset composed by i) a subset of the pool of candidates optimizing the MCLU criterion ($U^{\text{MCLU}}$) and ii)  the bounded support vectors sampled at the previous iteration, is iteratively partitioned in a binary way. The partitioning always considers the biggest current cluster found and continues until $q$ clusters not containing a bounded support vector are found. Once the $q$ clusters have been defined, a search among the candidates falling in these $q$ clusters is performed.

\begin{equation}
\hat{\x}^{\text{hMCS-i}} = \arg \min_{\x_i \in c_m } \Big\{f(\x_i)^{\text{MC}}   \Big\}\text{,}\quad m = [1, ..., q \vert n^{bSV}_{c_m} = 0], \x_i \in U^{\text{MCLU}}
\label{eq:hMCS}
\end{equation}

\section{Posterior probability based active learning}
\label{sec:poster}
The third class of methods uses the estimation of posterior probabilities of class membership (i.e. $p( y | \x)$) to rank the candidates. The posterior probability gives an idea of the confidence of the class assignment (which is usually done by maximizing it over all the possible classes): by considering the change on the overall posterior distribution or the per-class distribution for each candidate, these heuristics use these probability estimates to focus sampling in uncertain areas. This section details two heuristics, the KL-max and the Breaking ties.

\subsection{KL-max}
The first idea is to sample the pixels whose inclusion in the training set would maximize the changes in the posterior distribution. An application of these methods can be found in ~\cite{Roy01}, where the heuristic maximizes the Kullback-Leibler divergence between the distributions before and after adding the candidate. In remote sensing, a probabilistic method based on this strategy and using a Maximum Likelihood classifier can be found in~\cite{Raj08b}. In this setting, each candidate is removed from $U$ and it is included in the training set with the label maximizing its posterior probability. The Kullbach-Leibler divergence KL is then computed between the posterior distributions of the models with and without the candidate. After computing this measure for all candidates, the pixel maximizing the following is chosen:
{\small
\begin{align}
&\hat{\x}^{\text{KL-max}} =   \\\nonumber &= \arg \max_{\x_i \in U } \Bigg\{\sum_{\omega \in N} \frac{1}{(u-1)}  \text{KL}\Big(p^+(\omega|\x)\Big\vert\Big\vert p(\omega|\x)\Big)p(y^*_i = \omega|\x_i)\Bigg\}\label{eq:Kmax}
\end{align} }
where
{\small
\begin{align}
 \text{KL}\Big(p^+( \omega|\x)\Big\vert\Big\vert p( \omega|\x)\Big) = \sum_{\x_j \in U \setminus \x_i} p^+(y^*_j = \omega|\x_j)\log\frac{p^+(y^*_j = \omega|\x_j)}{p(y^*_j = \omega|\x_j)}
\end{align} }

and $p^+(\omega|\x)$ is the posterior distribution for class $\omega$ and pixel $\x$, estimated using the increased training set $X^+ = [X,(\x_i,y^*_i)]$, where $y_i^*$ is the class maximizing the posterior probability. Recently, the authors of~\cite{Jun08} extended this approach, proposing to use boosting to weight pixels that were previously selected, but were no longer relevant to the current classifier. These heuristics are useful when used with classifiers with small computational cost: since each iteration implies to train $u+1$ models, this type of heuristics is hardly applicable with computationally demanding methods as SVM. Moreover, a selection of batches of pixels is not possible.
\subsection{Breaking ties (BT)}
Another strategy, closer to the idea of EQB presented in Section \ref{sec:EQB}, consists of building a heuristic exploiting the conditional probability of predicting a given label $p(y_i^* = \omega|\x_i)$ for each candidate $\x_i \in U$. In this case, note that the predictions for the single candidates $y_i^* = \arg\max_{\omega \in N}f(\x_i,\omega)$ are used. Such estimates are provided by several methods, including probabilistic neural networks or maximum likelihood classifiers. A possibility to obtain posterior probabilities from SVM outputs\footnote{Even though they are not posterior probabilities from a Bayesian point of view} is to use Platt's estimation~\cite{Pla99}. 
In this case, the per-class posterior probability is assessed fitting a sigmoid function to the SVM decision function~\cite{Pla99}:
\begin{equation}
	p(y_i^* = \omega|\x_i) =\frac{1}{1 + e^{(Af(\x_i,\omega)+B)}}
\end{equation}
\mitch{where $\omega = \{1,... ,N\}$ is the class considered and $f(\x_i,\omega)$ is the binary decision function of the SVM without the sign operator (see Eq.~\eqref{eq:decfunct}).}{} where $A$ and $B$ are parameters to be estimated (for details, see~\cite{Pla99}). 
Once the posterior probabilities are obtained, it is possible to assess the uncertainty of the class membership for each candidate in a direct way. In this case the heuristic choses candidates showing a near uniform probability of belonging to each class, i.e. $p( y_i^* = \omega | \x_i) = 1/N, \  \forall \omega \in N$.

The `Breaking ties' (BT) heuristic for a binary problem relies on the smallest difference of the posterior probabilities for each sample \cite{Luo05}. In a multi-class setting, this reciprocity can still be confirmed and used, since independently from the number of classes $N$, the difference between the two highest probabilities can be indicative of the way an example is handled by the classifier\mitch{(see a similar interpretation for the MCLU heuristic in Section~\ref{sec:MCLU})}{}. When the two highest probabilities are close (``on a tie''),  the classifier's confidence is the lowest. The BT heuristic can thus  be formulated as:

\begin{equation}
\hat{\x}^{\text{BT}} = \arg \min_{\x_i \in U } \Big\{\max_{\omega \in N} \big\{p(y_i^* = \omega|\x)\big\} - \max_{\omega \in \mitch{}{N}\backslash \omega^+} \big\{p(y_i^* = \omega|\x)\big\}\Big\}
\label{eq:BT}
\end{equation}
where $\omega^+$ is the class showing maximal probability, i.e. the argument of the first term of Eq.~\eqref{eq:BT}. By comparing Eq.~\eqref{eq:MCLU} with Eq.~\eqref{eq:BT}, it is clear that the link between BT and the MCLU heuristic when using SVM classifiers (see Section~\ref{sec:MCLU}) is really strong.

\section{Datasets and setup}\label{sec:data}
This Section details the datasets considered and the setup of the experiments performed.

\subsection{Datasets}
\begin{figure}[!b]
\centering
	\begin{tabular}{cc}
	\multicolumn{2}{c}{\includegraphics[width=8.4cm]{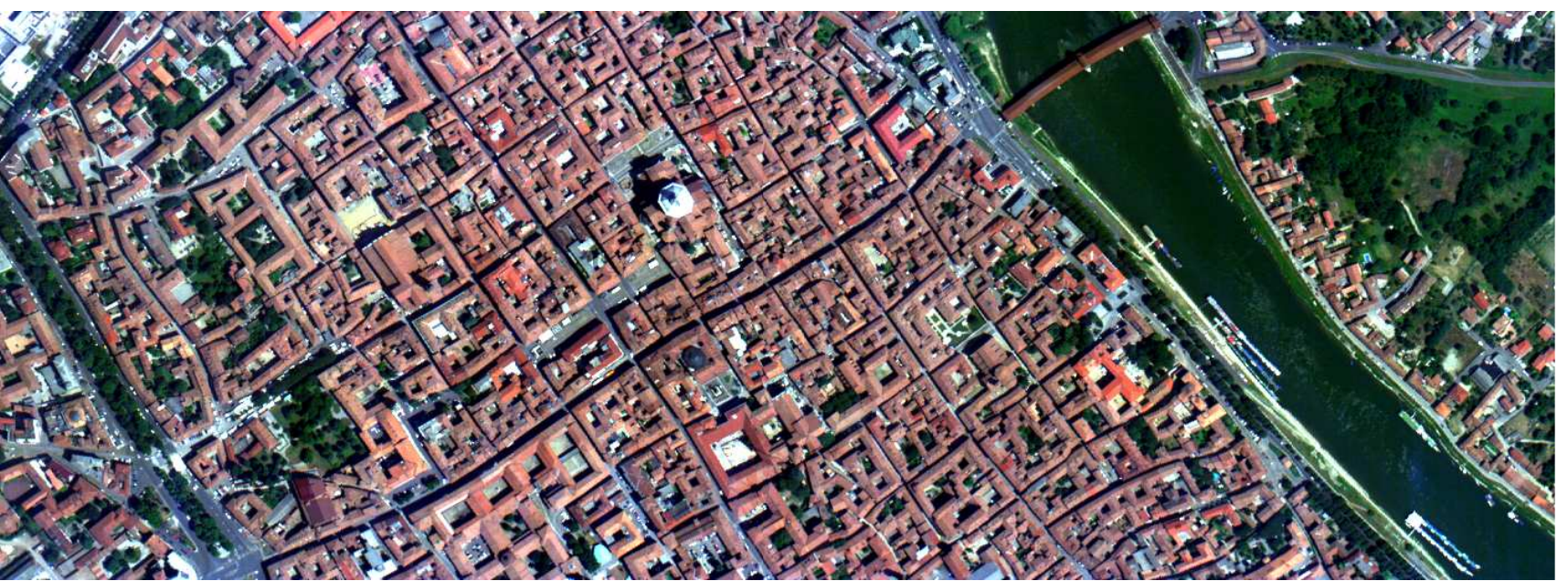}}\\
	\multicolumn{2}{c}{\includegraphics[width=8.4cm]{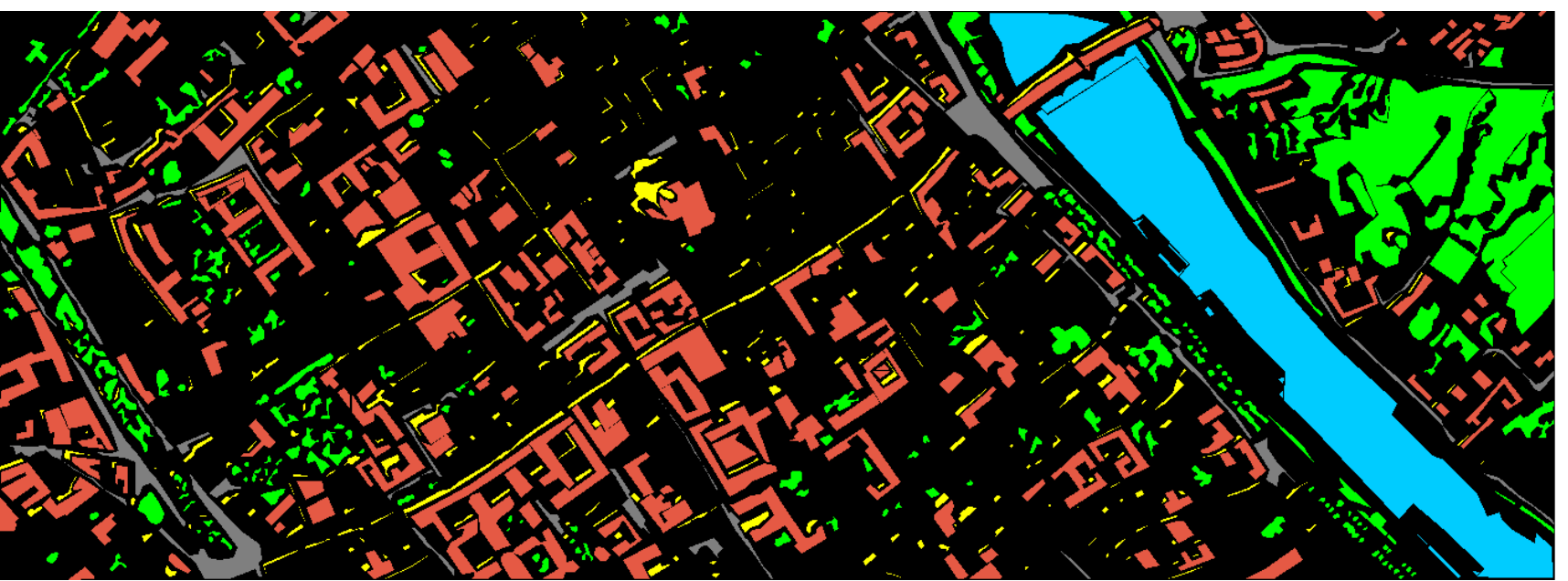}}\\
	\multicolumn{2}{c}{ROSIS Pavia}\\

	\includegraphics[height=4cm]{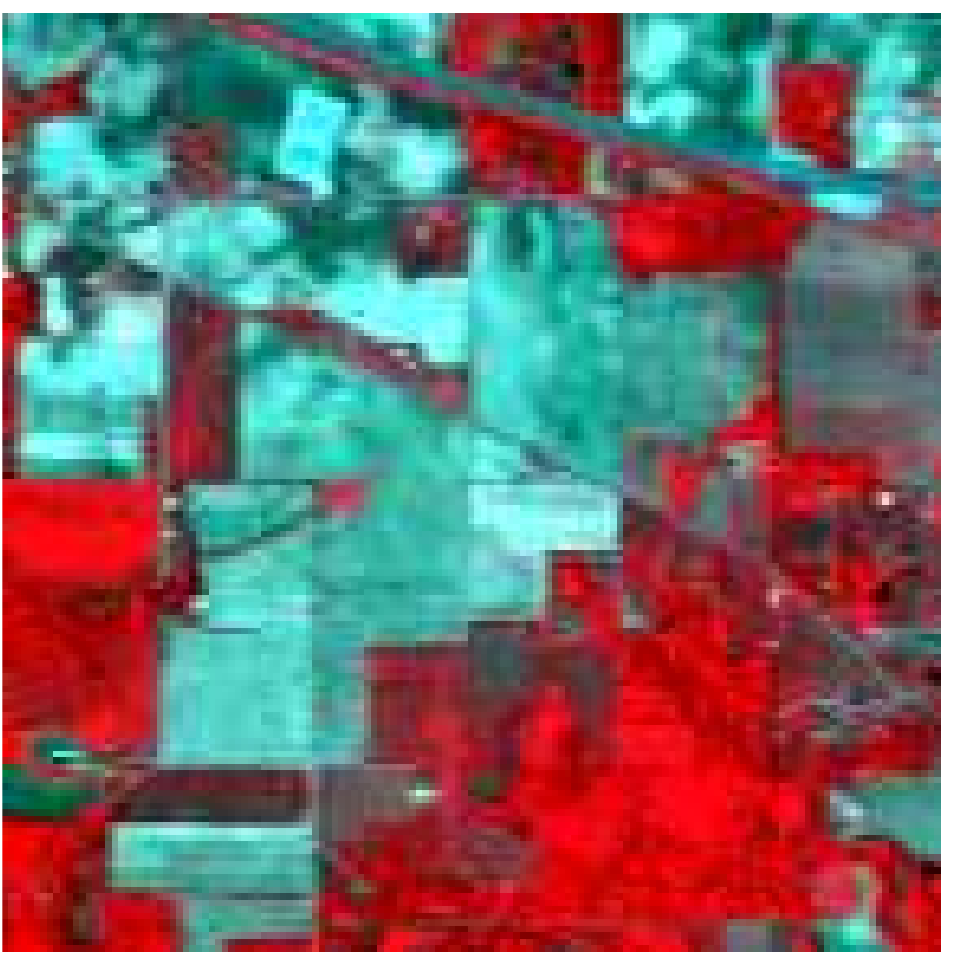}&
	\includegraphics[height=4cm]{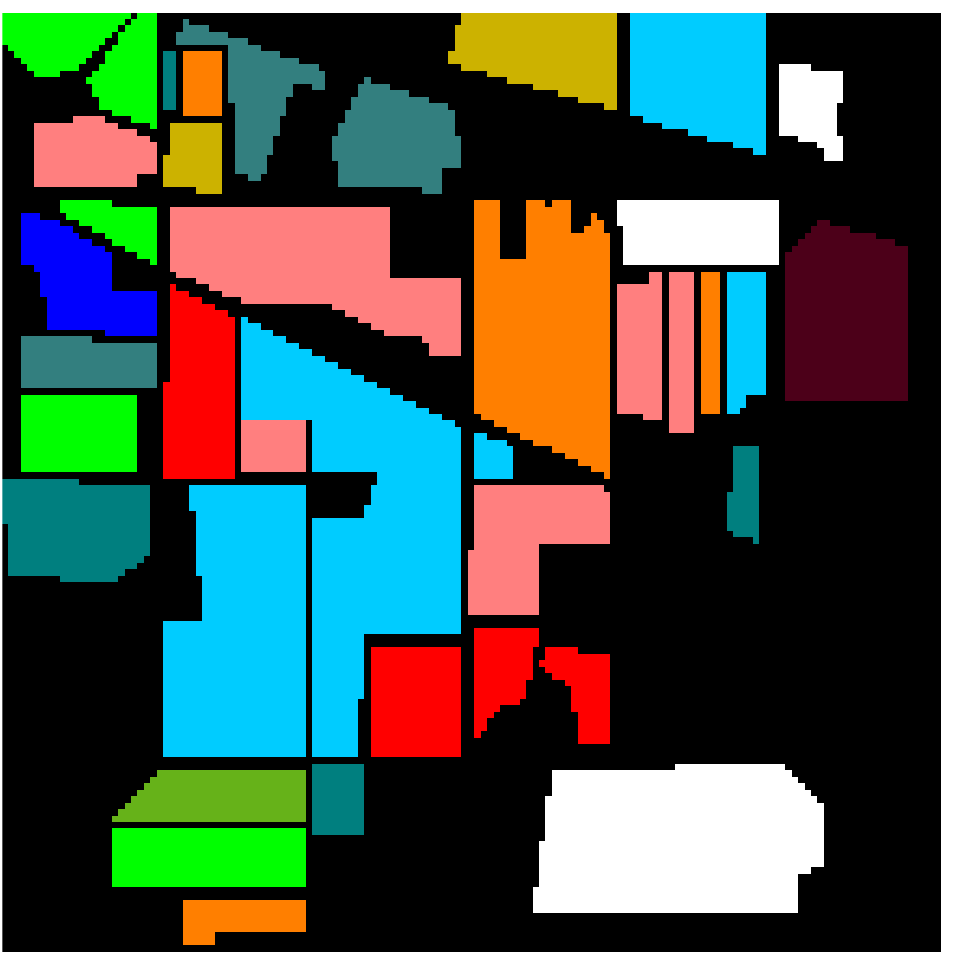}\\
	\multicolumn{2}{c}{AVIRIS Indian Pines}\\
	
\includegraphics[height=4cm]{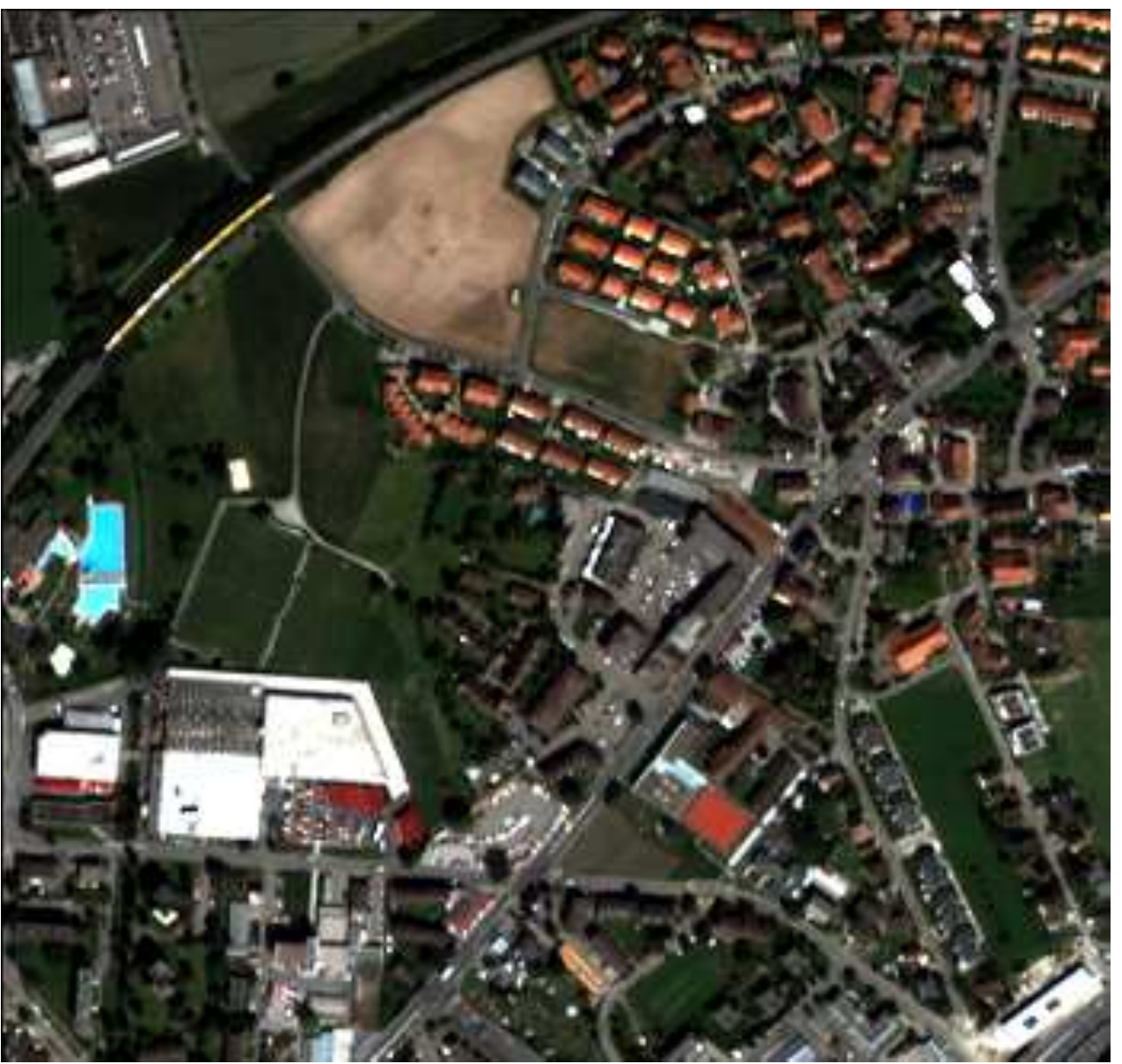}&
\includegraphics[height=4cm]{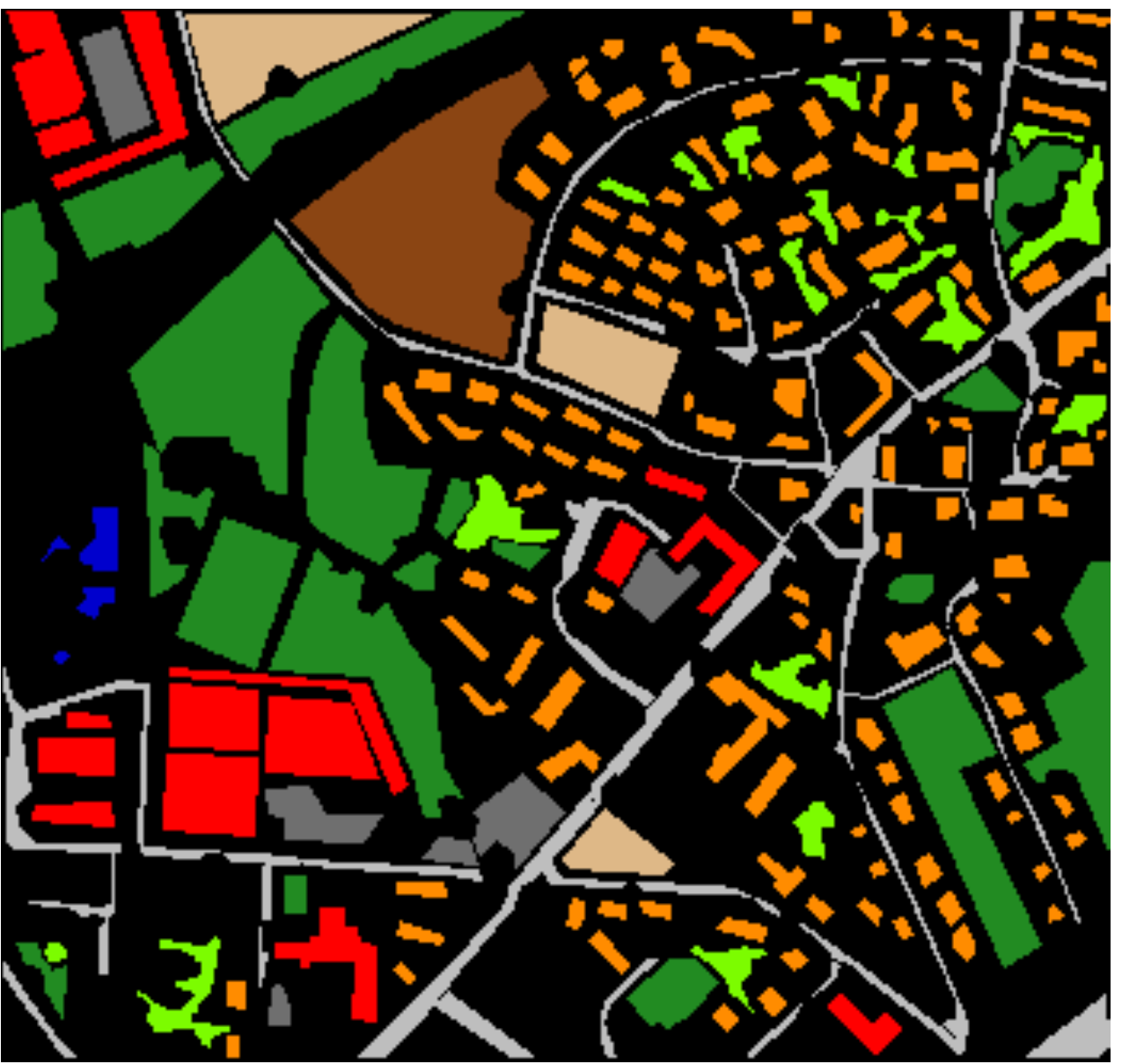}\\
	\multicolumn{2}{c}{QuickBird Zurich}\\
	\end{tabular}
	\caption{Images considered in the experiments: (top) ROSIS image of the city of Pavia, Italy (bands $[56-31-6]$ and corresponding ground survey); (middle) AVIRIS Indian Pines hyperspectral data (bands $[40-30-20]$ and corresponding ground survey); (bottom) QuickBird multispectral image of a suburb of the city of Zurich, Switzerland (bands $[3-2-1]$ and corresponding ground survey).}
	\label{fig:data}
\end{figure}

\begin{figure}
\centering
\begin{tabular}{ccc}
\includegraphics[width=4cm]{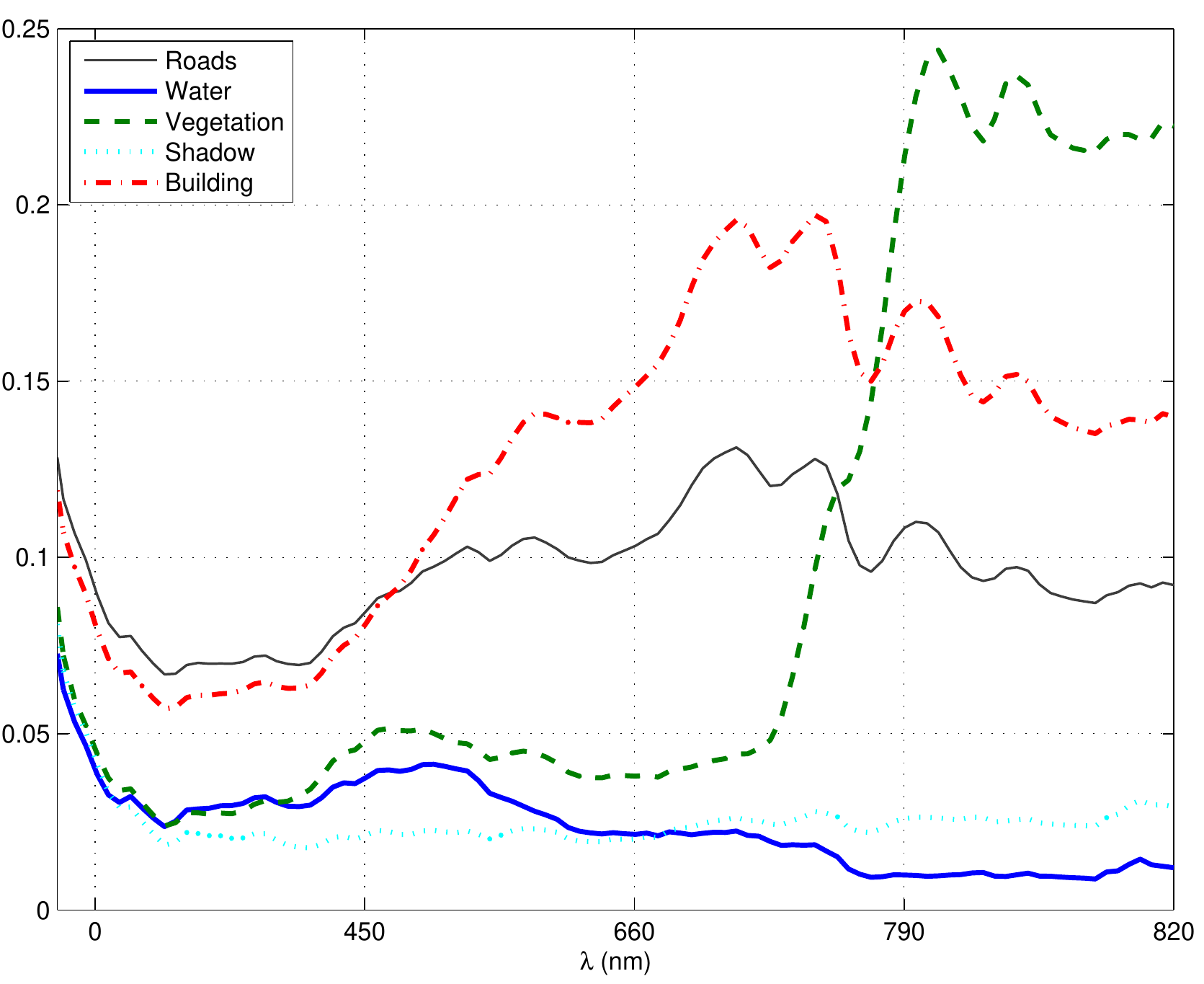}&
\includegraphics[width=4cm]{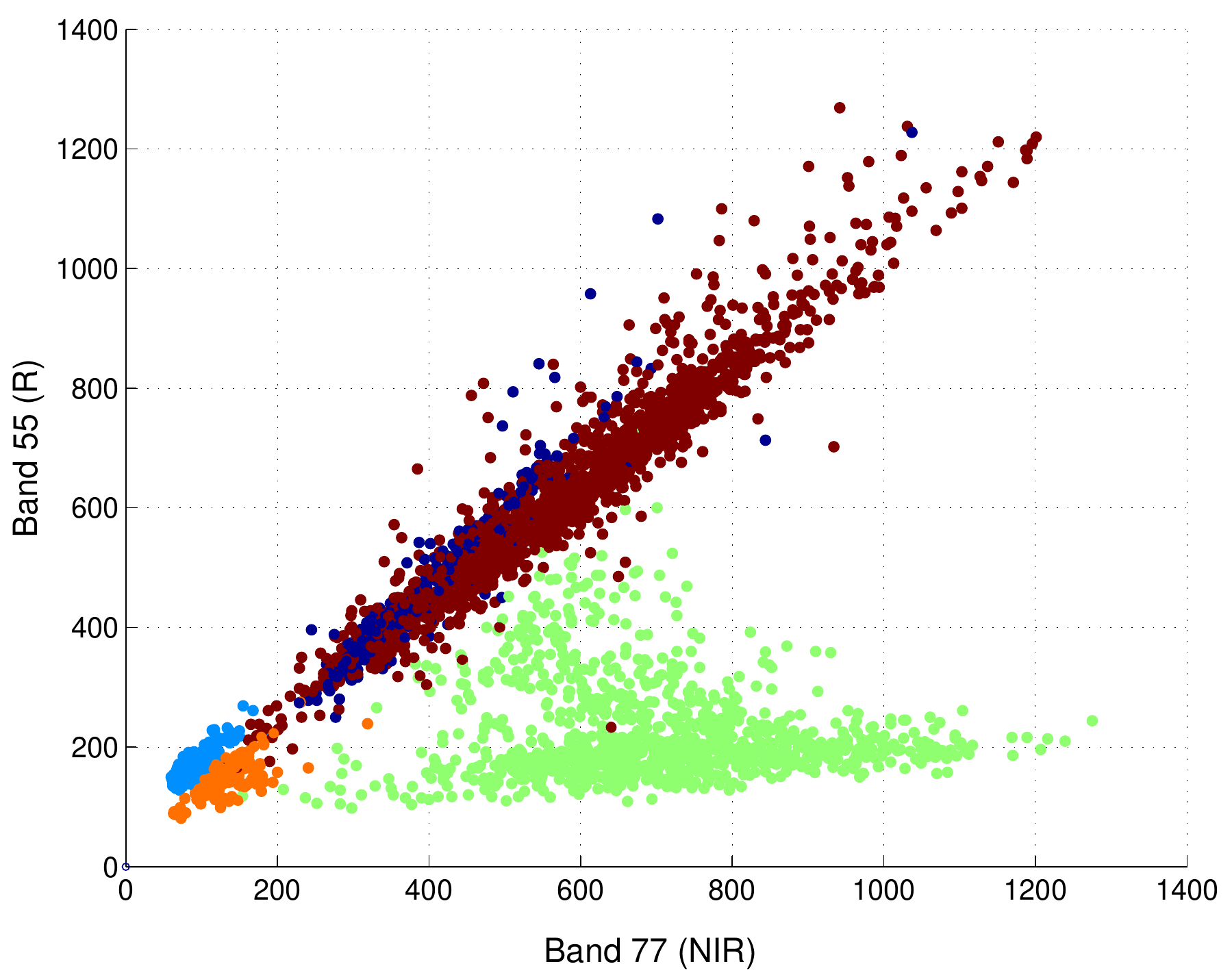}\\
(a)&(b)\\
\includegraphics[width=4cm]{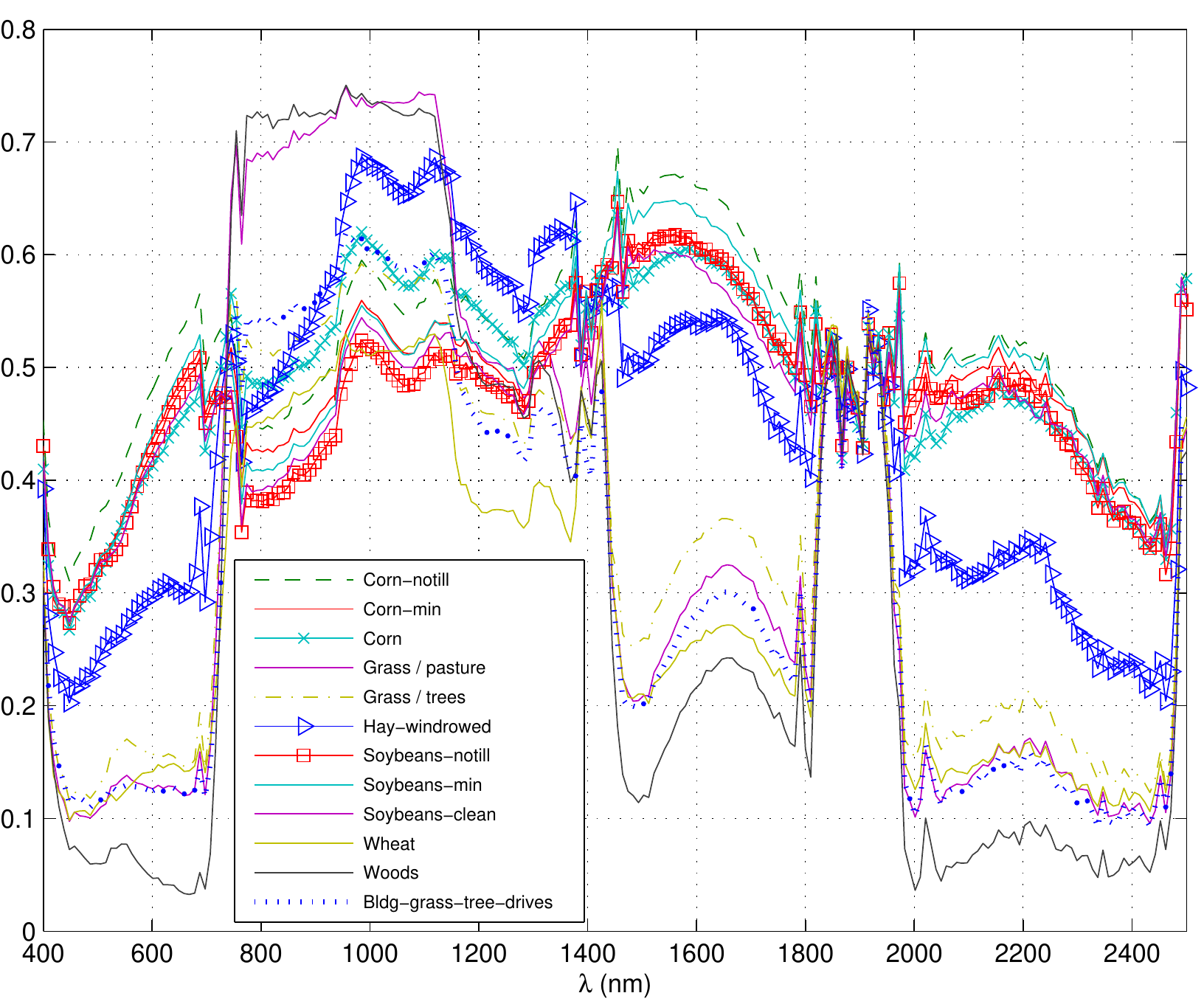}&
\includegraphics[width=4cm]{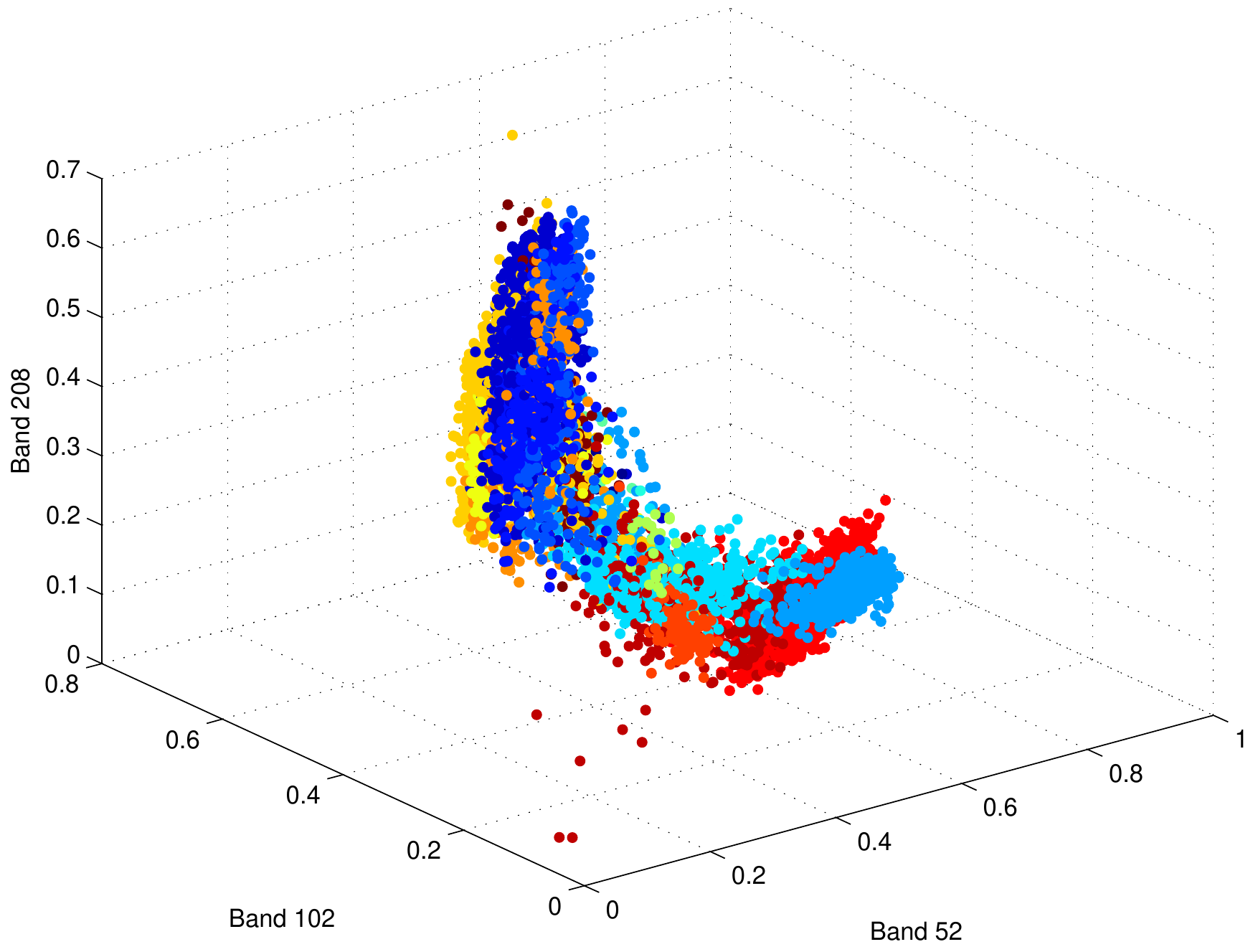}
\\
(c)&(d)\\
\includegraphics[width=4cm]{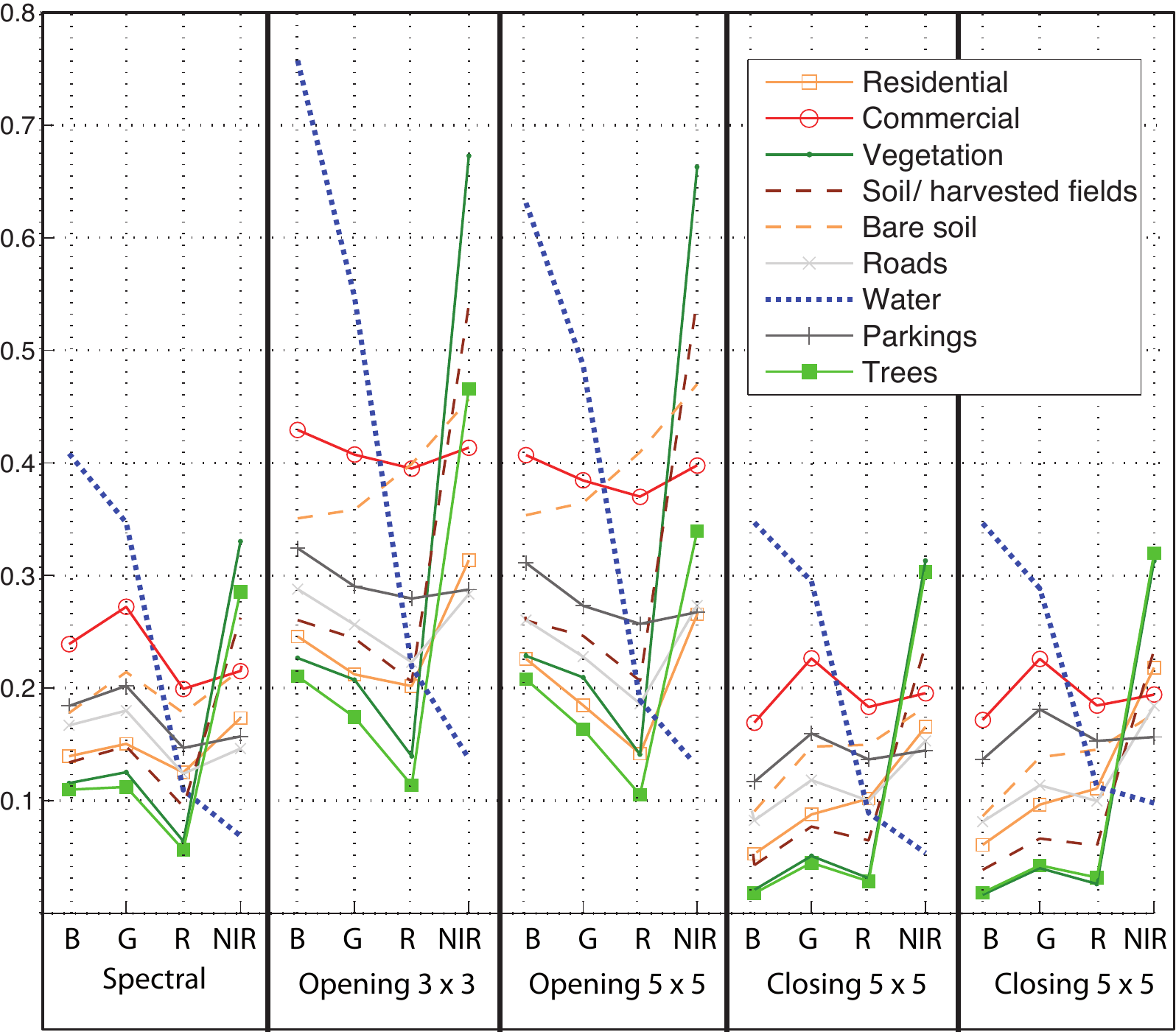}&
\includegraphics[width=4cm]{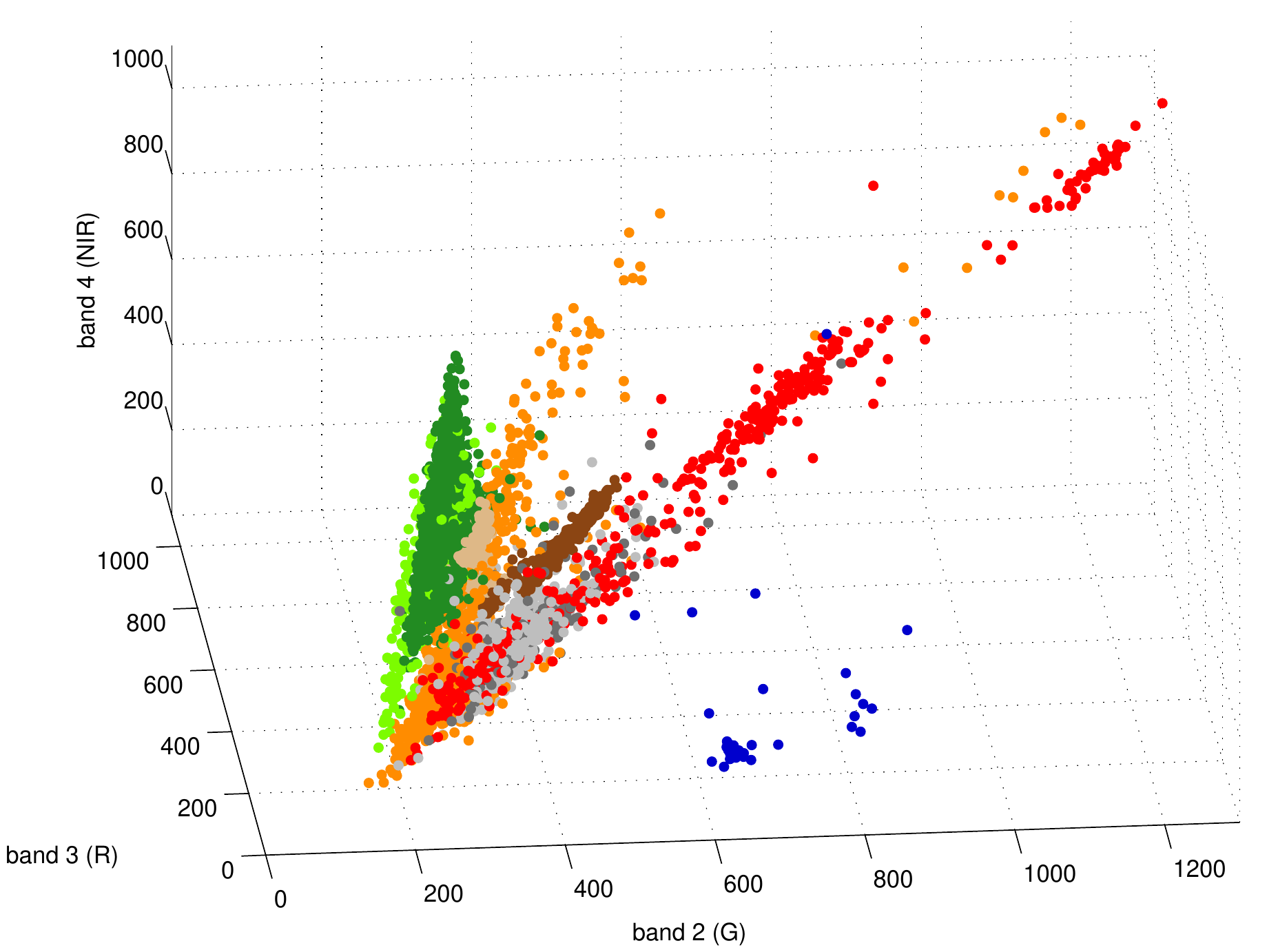}
\\
(e) & (f)\\
\end{tabular}
\caption{Data distribution of the three images considered. First row: ROSIS image of Pavia: (a) mean spectral profiles; (b)  example of data manifold in bands 55 (Red) and 77 (Near infrared). Middle row: AVIRIS Indian Pines: (c) mean spectral profiles; (d) example of data manifold in bands 52, 102 and 208. Bottom row: Zurich QuickBird: (e) mean spectral profiles; (f) data manifold in bands 2 (G), 3 (R) and 4 (NIR).}

\label{fig:Distr}
\end{figure}

Active heuristics have been tested on three challenging remote sensing classification scenarios (Fig.~\ref{fig:data}), whose data distributions are detailed in Fig.~\ref{fig:Distr}. 

\subsubsection{Hyperspectral VHR} the two top rows of Fig.~\ref{fig:data} show a hyperspectral 1.3m spatial resolution image of the city of Pavia (Italy) taken by the airborne ROSIS-03 optical sensor~\cite{Lic09}. The image consists of 102 spectral bands of size ($1400 \times 512$) pixels with a spectral coverage ranging from 0.43 to 0.86 $\mu$m. $5$ classes of interest (Buildings, Roads, Water, Vegetation and Shadows) have been selected and a labeled dataset of 206`009 pixels has been extracted by visual inspection. Among the available pixels, 20`000 have been used for the training set $X$ and candidate set $U$. Ten independent experiments have been performed, starting with $5 \times 5 = 25$ labeled pixels ($5$ per class) in $X$ and the remaining pixels in $U$. When using LDA,  150 pixels ($30$  per class) have been included in the starting set. The higher number of starting training pixels used for LDA is justified by the requirements of the model ($n$ must be greater than the dimensionality of the data). In each experiment, 80`000 randomly selected pixels have been used to test the generalization capabilities of the heuristics.

The data distribution of the five classes is illustrated in the first row of Fig.~\ref{fig:Distr}: from the mean spectra (Fig.~\ref{fig:Distr}(a)) the classes are well distinguished and separable with the sole spectral information and the resulting data manifold (Fig.~\ref{fig:Distr}(b)) shows a data distribution which can be handled by most linear and non linear models.

\subsubsection{Hyperspectral MR} the second dataset, illustrated in the second row of Fig.~\ref{fig:data}, is a 220-bands AVIRIS image taken over Indiana's Indian Pine test site in June 1992~\cite{Jac01}. The image is $145 \times 145$ pixels, contains 16  classes representing different crops, and a total of 10`366 labeled pixels. This image is a classical benchmark to validate model accuracy and constitutes a very challenging classification problem because of the strong mixture of the class signatures. Twenty water absorption channels were removed prior to analysis
. In the experiments, classes with less than 100 labeled pixels were removed, resulting thus in a 12 classes classification problem with 10`171 labeled pixels (see the ground truth pixels in Fig.~\ref{fig:data}). Among the available labeled pixels, 7`000 were used for the $X$ and $U$ sets. Each experiment starts with $5 \times 12 =60$ pixels (5 per class). As for the previous image, the remaining 3`171 pixels have been used to test the generalization capabilities.

Visualization of the spectral mean profiles and of the data manifold (second row of Fig.~\ref{fig:Distr}) illustrates a completely different situation with respect to the previous image: high nonlinearity and strongly overlapping classes characterize this dataset. Therefore, linear classifiers do not perform well on this dataset and will not be considered in the experiments.

\subsubsection{Multispectral VHR}
The third image, illustrated in the last row of Fig.~\ref{fig:data}, is a 4-bands QuickBird scene of a suburb of the city of Zurich (Switzerland) taken in 2002. The image is $329 \times 347$ pixels with a spatial resolution of 2.4m. Nine classes of interest have been extracted by careful visual inspection, namely Residential buildings, Commercial buildings, Trees, Vegetation, Harvested fields, Bare soil, Roads, Parking lots and Water. Since some of the classes to be separated are of landuse and have very similar responses in the spectral domain (see, for instance, the residential and commercial buildings, or roads and parking lots), 16 contextual bands, extracted using opening (8) and closing (8) morphological operators (see~\cite{Tui09a}), have been added to the four spectral bands, resulting in a 20-dimensional dataset. As for the Pavia dataset, 20`000 pixels have been extracted for the $X$ and $U$ sets. Each experiment starts with $5 \times 9 =45$ pixels ($5$ per class). 
The complexity of this third dataset is confirmed by both the spectra and the manifold illustrated in the bottom row of Fig.~\ref{fig:Distr}. Strong overlaps between the asphalt and the soil classes are observed, which is also confirmed by the similarity between the spectral profiles. However, the spatial features added improve the differentiation of the classes (see, for instance, the opening features for the vegetation classes and the closing features for the asphalt classes). 

\subsection{Experimental setup} 
In the experiments, SVM classifiers with RBF kernel and LDA classifiers have been considered for the experiments.  When using SVM, free parameters have been optimized by 5-fold cross validation optimizing an accuracy criterion. The active learning algorithms have been run in two settings, adding $N+5$ and $N+20$ pixels per iteration. To reach convergence, $70$ ($40$ in the case $N + 20$) iterations have been executed for the first image, $100$ ($50$) for the second and $80$ ($50$) for the third. $n$EQB has been run with committees of $7$ models using $75\%$ of the available training data. For the experiments using LDA, $40$ (20) iterations have been performed and $n$EQB using $12$ models and $85\%$ of the data have been used.

An upper bound on the classification accuracy has been computed by considering a model trained on the whole $X \cup U$ set (`Standard SVM/LDA' hereafter). The lower bound on performance has been considered by assessing a model using an increasing training set, randomly adding the same number of pixels at each epoch (`Random Sampling' hereafter). 

Each heuristic has been run ten times with different initial training sets. All the graphics report mean and standard deviation of the ten experiments.

\section{Numerical results}
\label{sec:res}
In this section, some of the heuristics presented are compared on the three datasets presented above. 
The experiments do not aim at defining which heuristic is best, since they respond unequally well to different data architectures. Rather, it attempts to illustrate the strengths and weaknesses of the different families of methods and to help the user in selecting the methodology that will perform best depending on the characteristics of the problem at hand. 
The heuristics studied are the following: $n$EQB, MS, MCLU, MCLU-ABD and BT. Their comparison with Random sampling (RS), the base learner (Standard SVM/LDA) and between each other will show the main differences among the active learning architectures presented. 

\begin{figure}[!t]
\begin{tabular}{ccc}
& $N+5$ & $N+20$\\
\rotatebox{90}{ Pavia ROSIS}
&
\includegraphics[width=3.8cm]{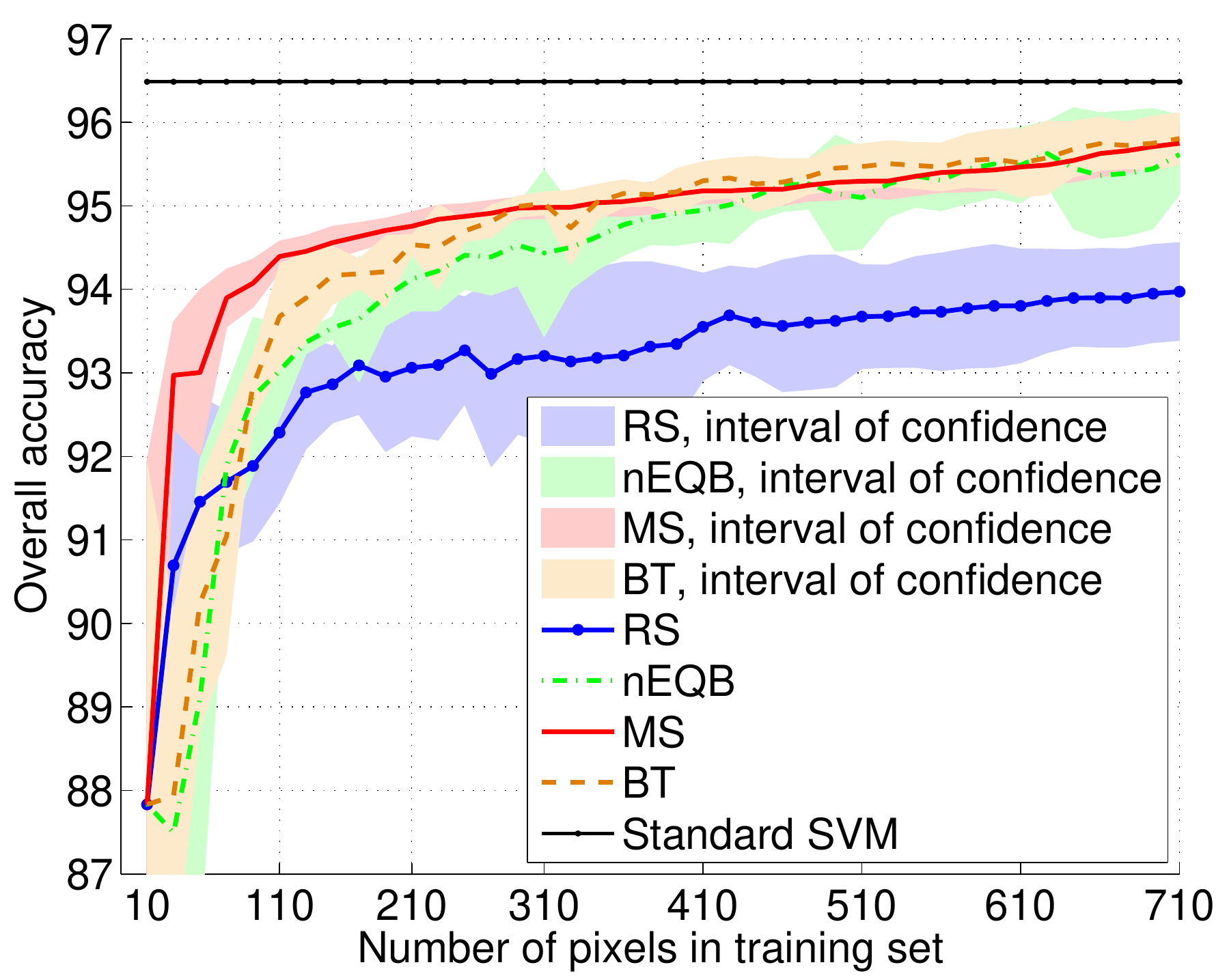}
&
\includegraphics[width=3.8cm]{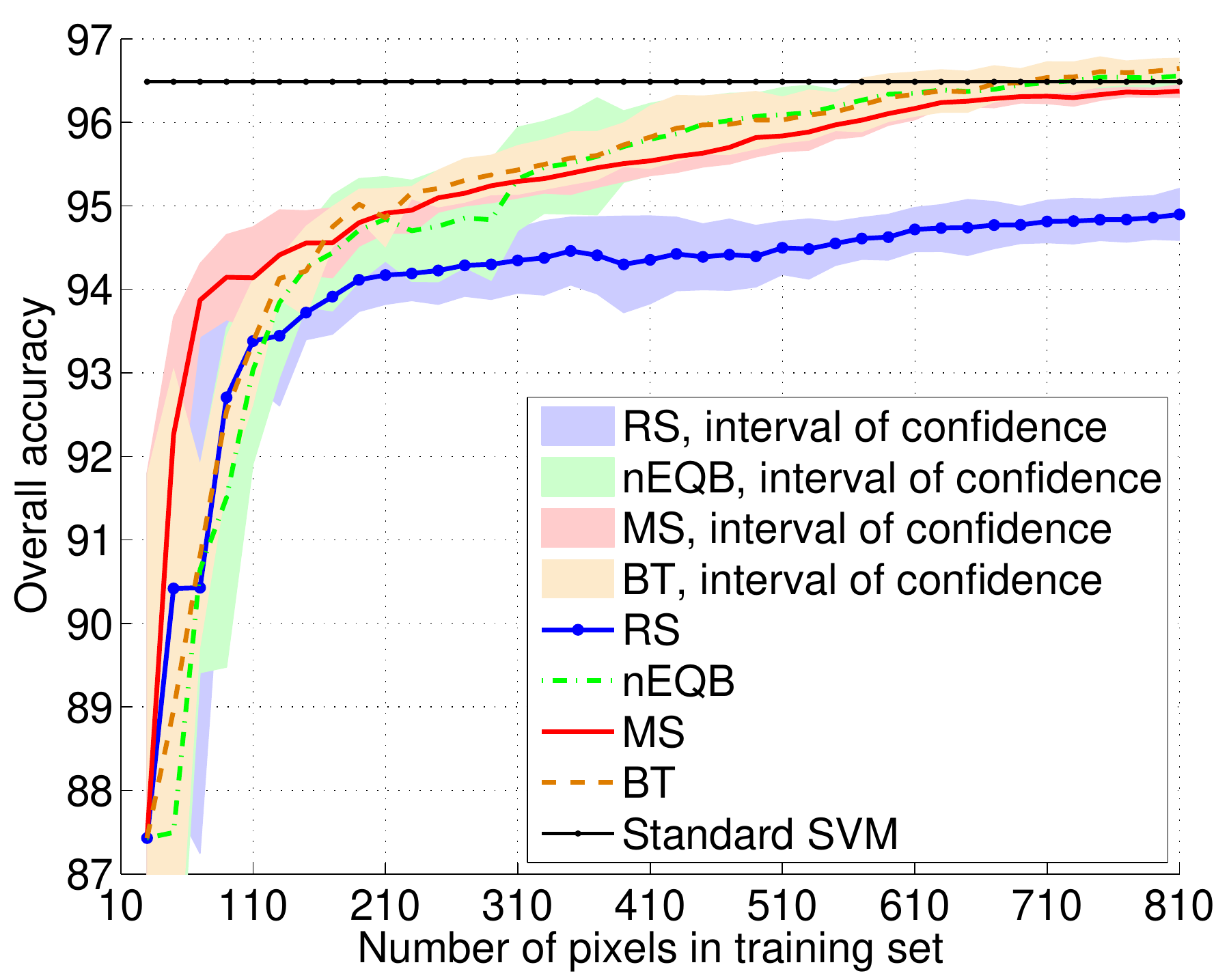}\\
&(a)&(d)\\

\rotatebox{90}{ Indian Pines AVIRIS}
&
\includegraphics[width=3.8cm]{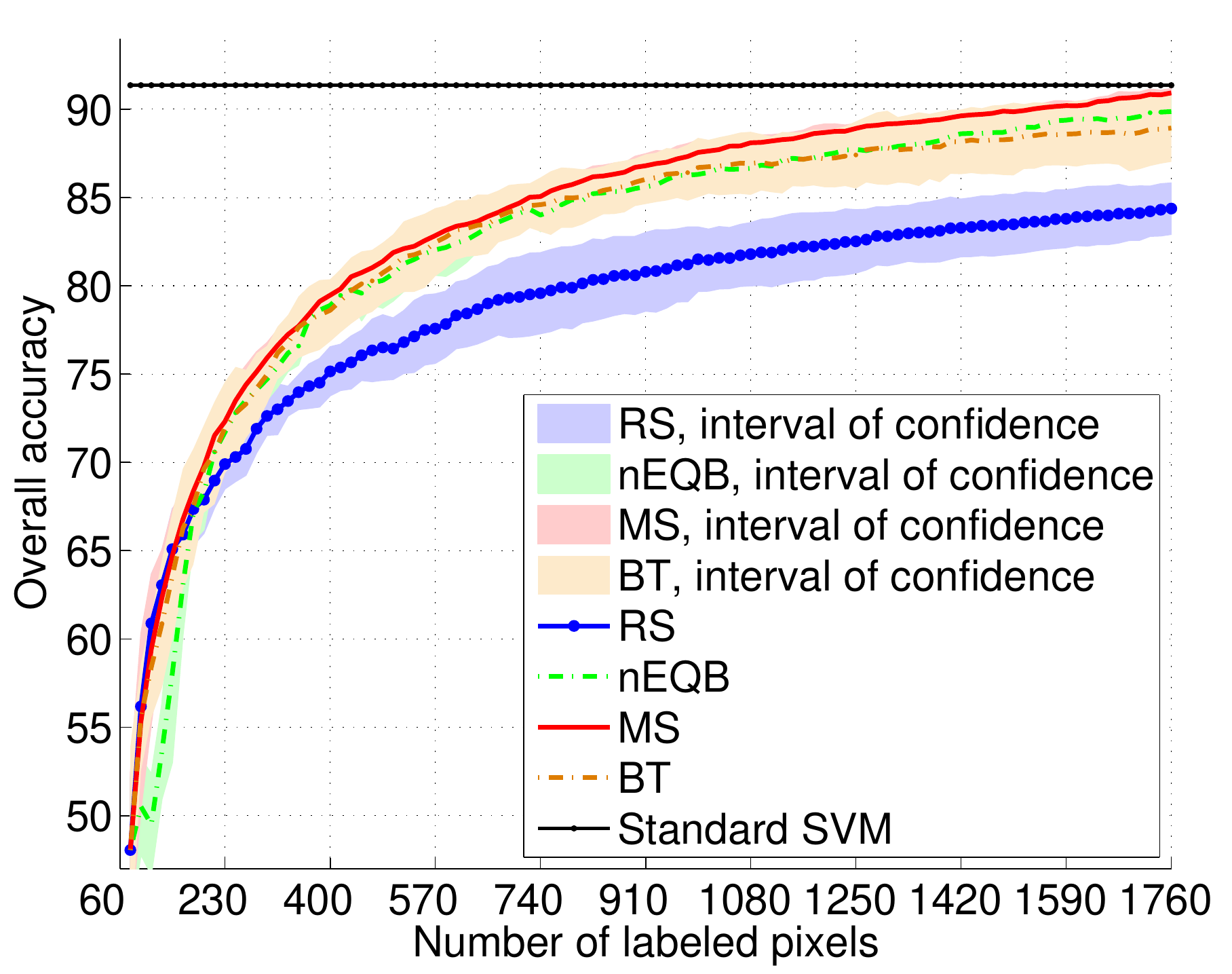}
&
\includegraphics[width=3.8cm]{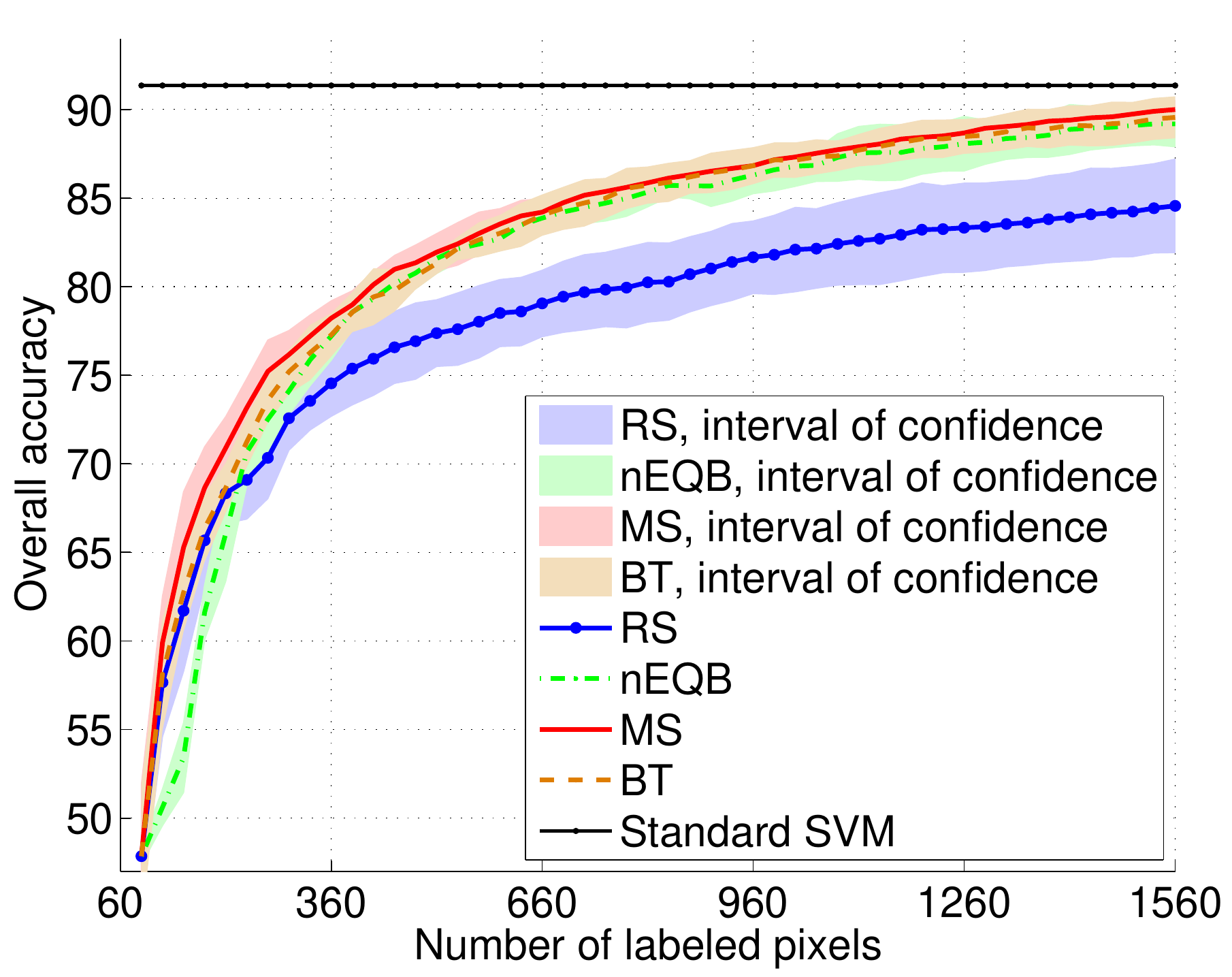}
\\
&(b)&(e)\\

\rotatebox{90}{ Zurich QuickBird}
&
\includegraphics[width=3.8cm]{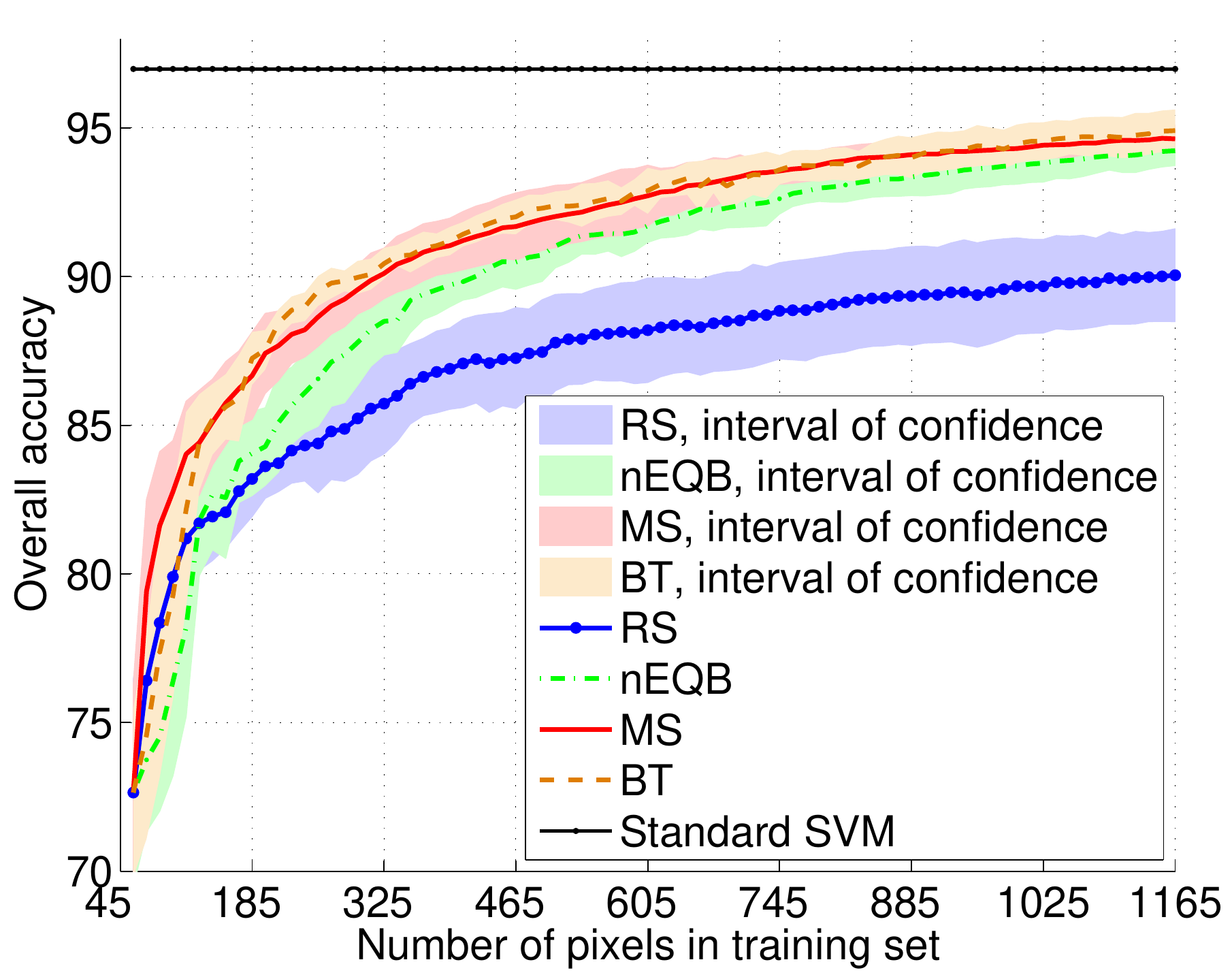}
&
\includegraphics[width=3.8cm]{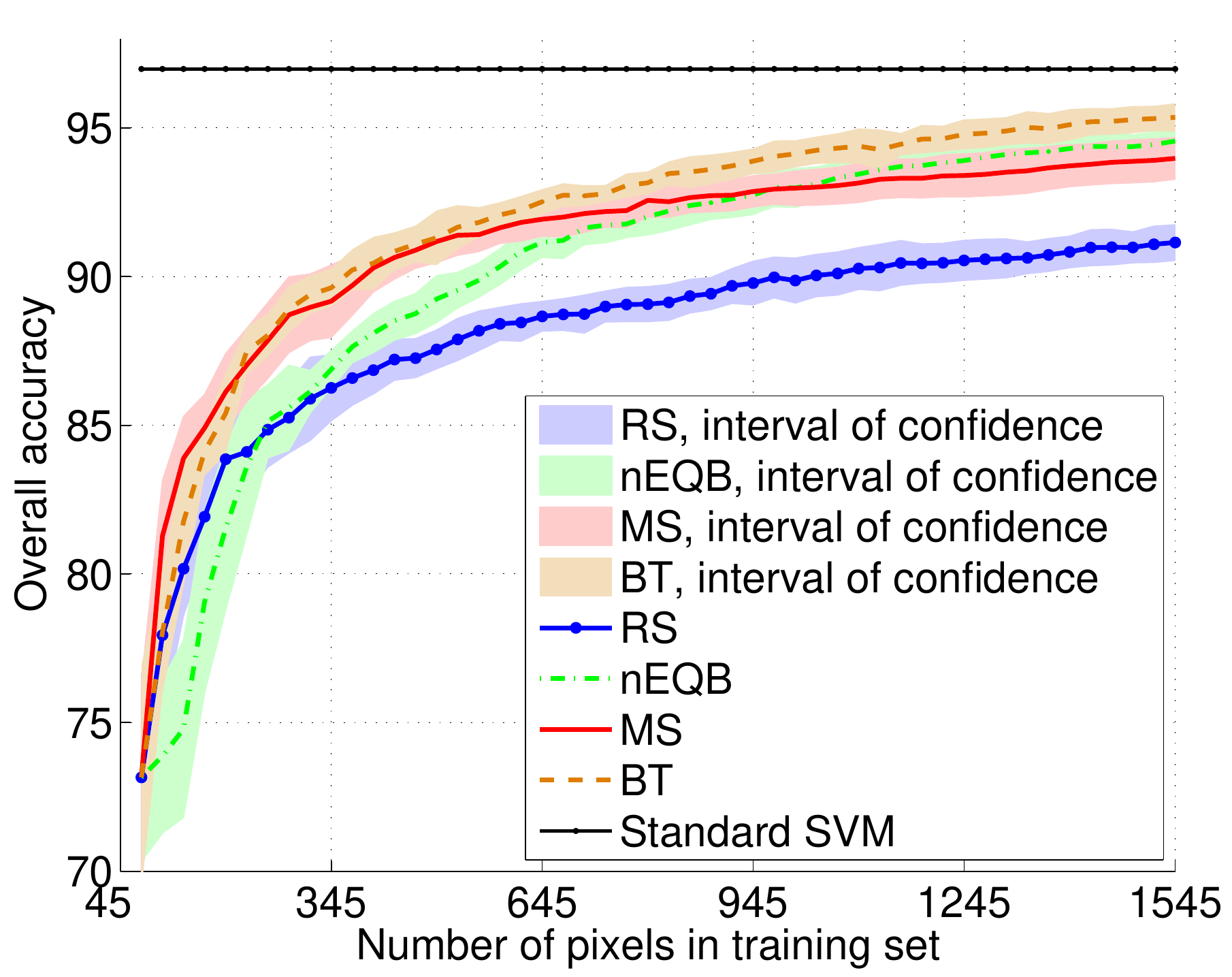}
\\
&(c)&(f)\\

\end{tabular}
\caption{The three families of heuristics trained with SVMs (RS = Random Sampling).}
\label{fig:comp}
\end{figure}

Figure~\ref{fig:comp} compares a heuristic for each family presented, when using SVM classifiers. In general, MS performs better than the two other families. This is expected, since MS ranks the candidates directly using the SVM decision function without further estimations: the slightly inferior performances of the $n$EQB and the BT algorithms are mainly due to the small size of the initial training set, which does not allow these criteria to perform optimally. $n$EQB uses too small bags of training pixels and BT cannot estimate the posterior probabilities correctly, because the fit of Platt's probabilities is dependent on the number of samples used. As a consequence, the  performances of these two heuristics in the first iterations is similar to random sampling, a behavior already observed in~\cite{Tui09}. Summarizing, when using SVMs, the most logical choice among the families seems to be a large margin heuristic.

\begin{figure}[!b]
\begin{tabular}{ccc}
& $N+5$ & $N+20$\\
\rotatebox{90}{ Pavia ROSIS}
&
\includegraphics[width=3.8cm]{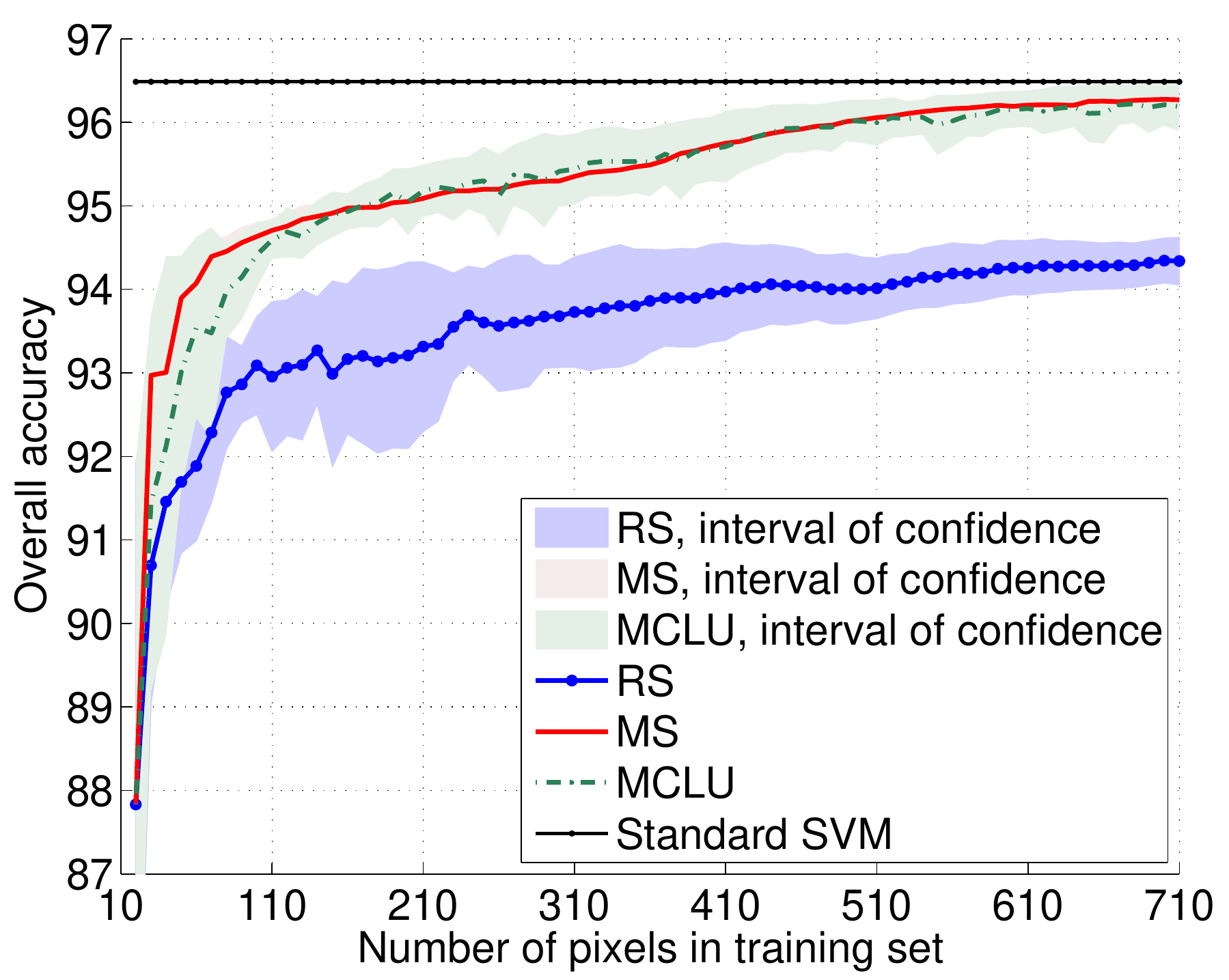}
&
\includegraphics[width=3.8cm]{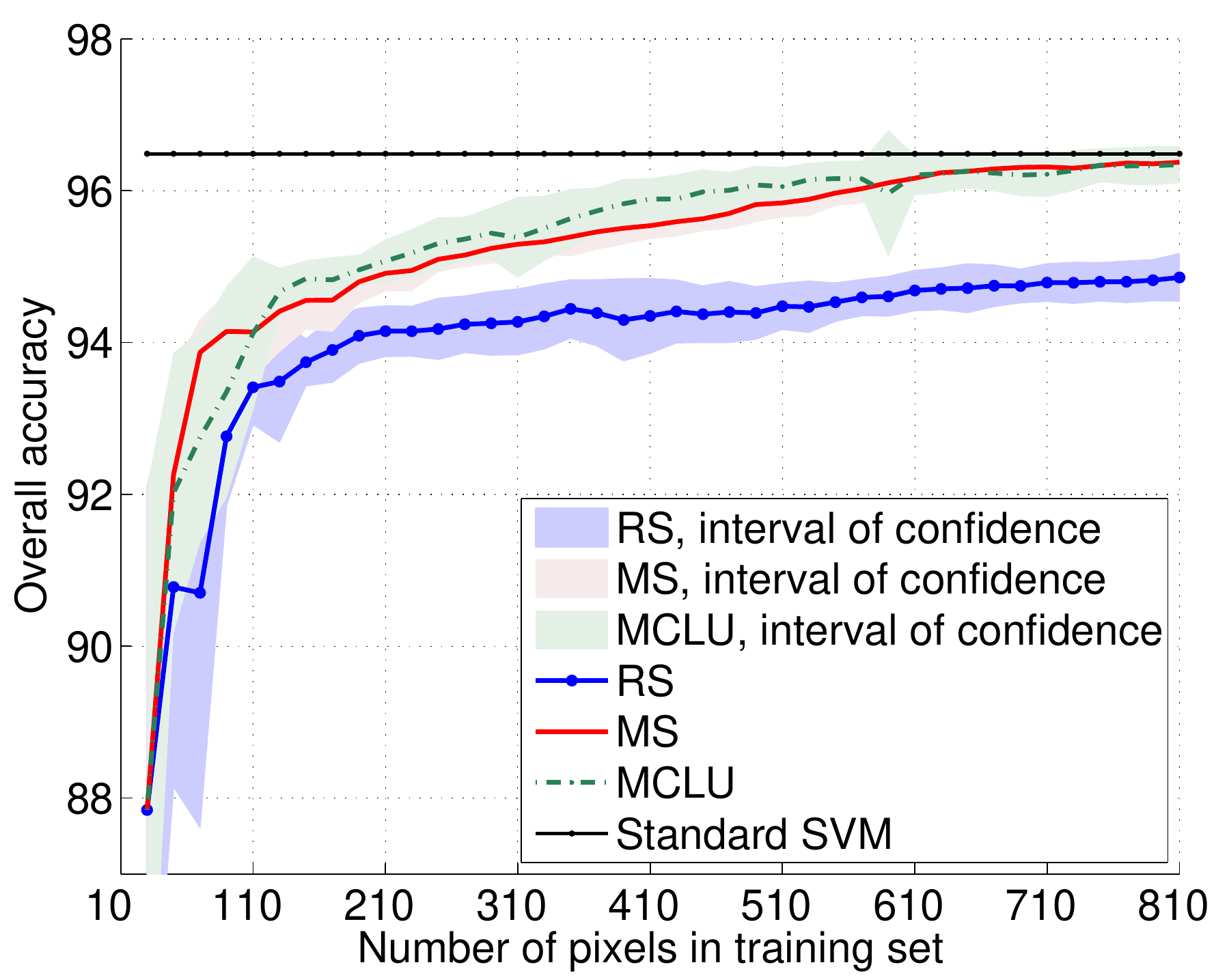} \\
&(a)&(d)\\

\rotatebox{90}{ Indian Pines AVIRIS}
&
\includegraphics[width=3.8cm]{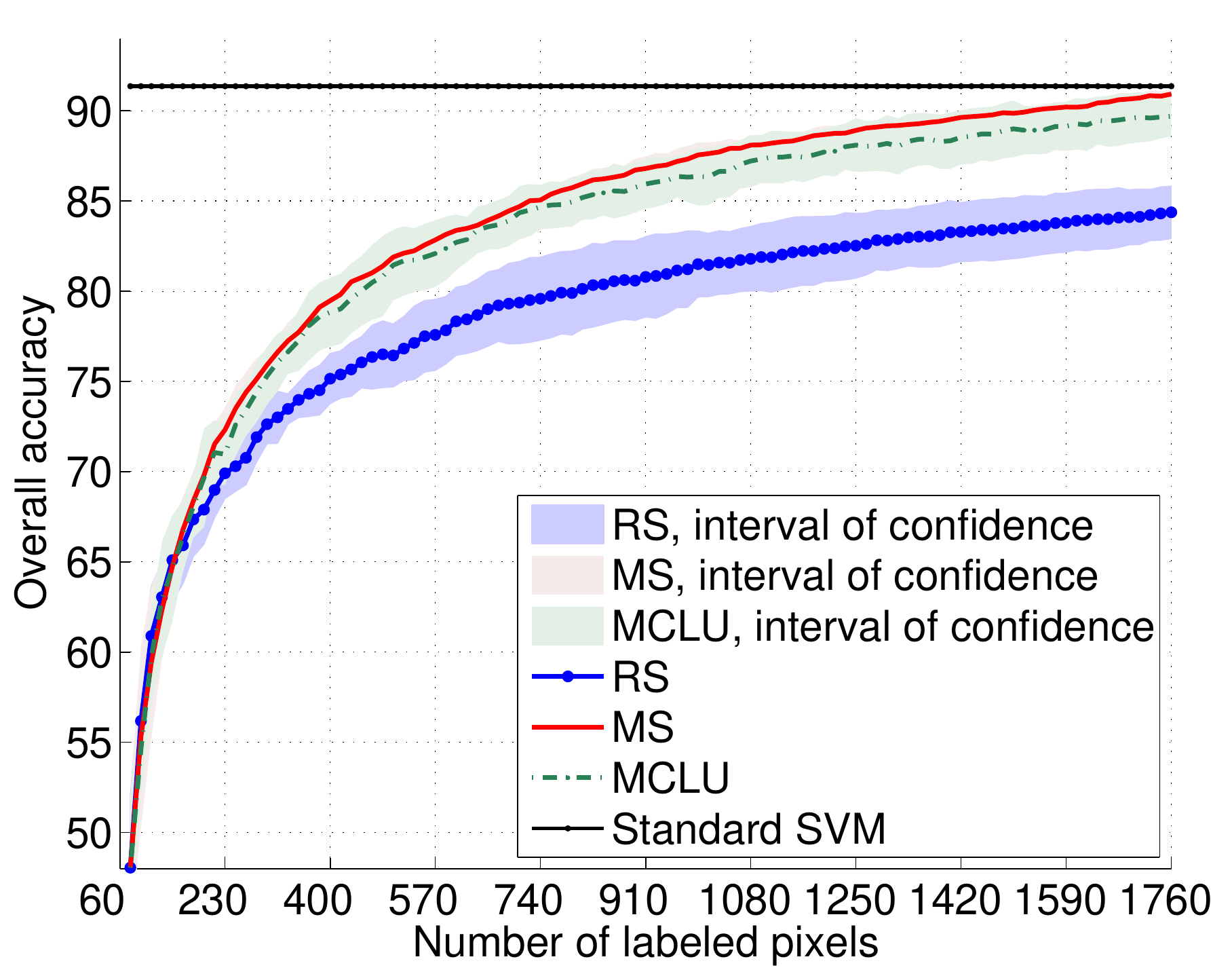}
&
\includegraphics[width=3.8cm]{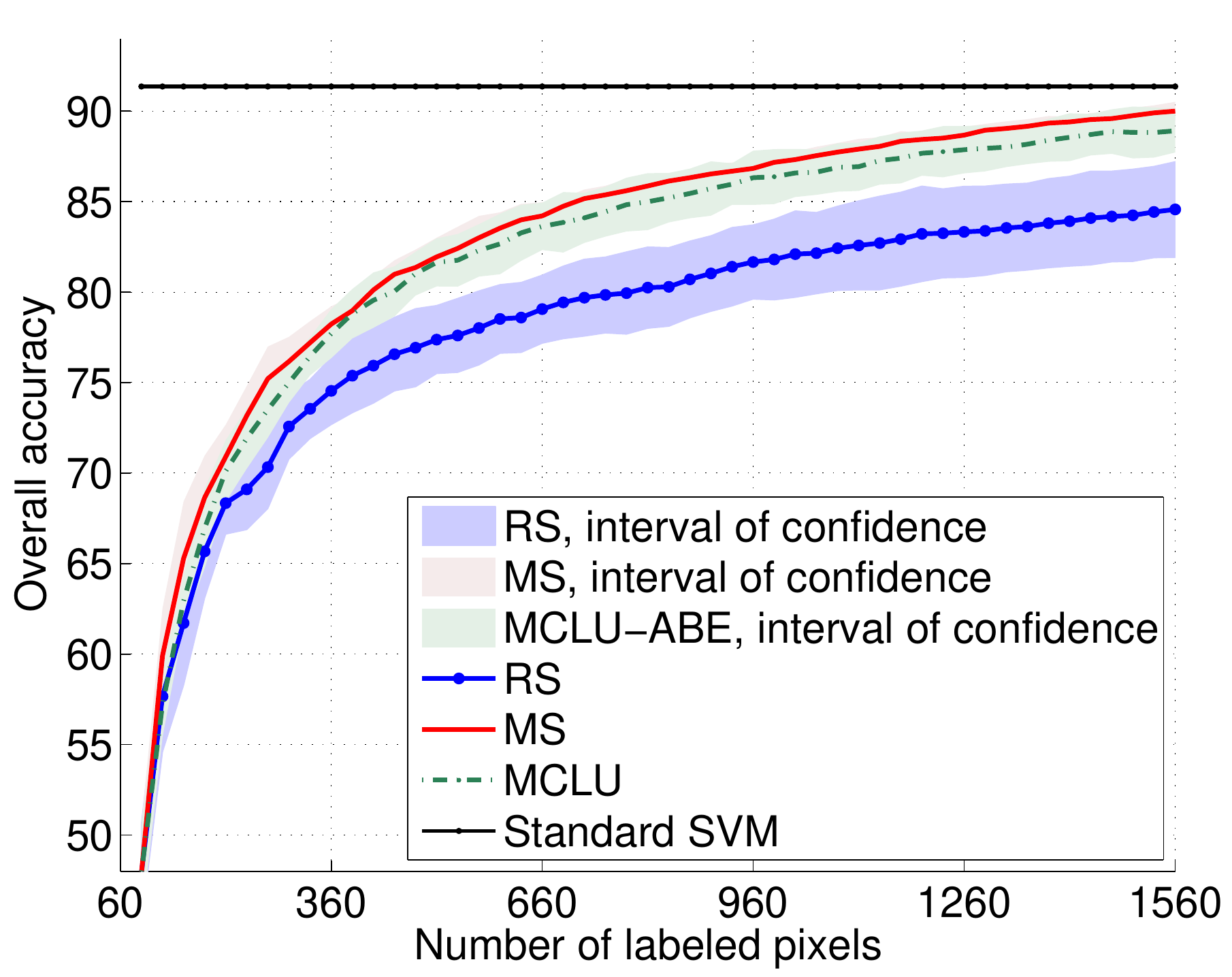} \\
&(b)&(e)\\

\rotatebox{90}{ Zurich QuickBird}
&
\includegraphics[width=3.8cm]{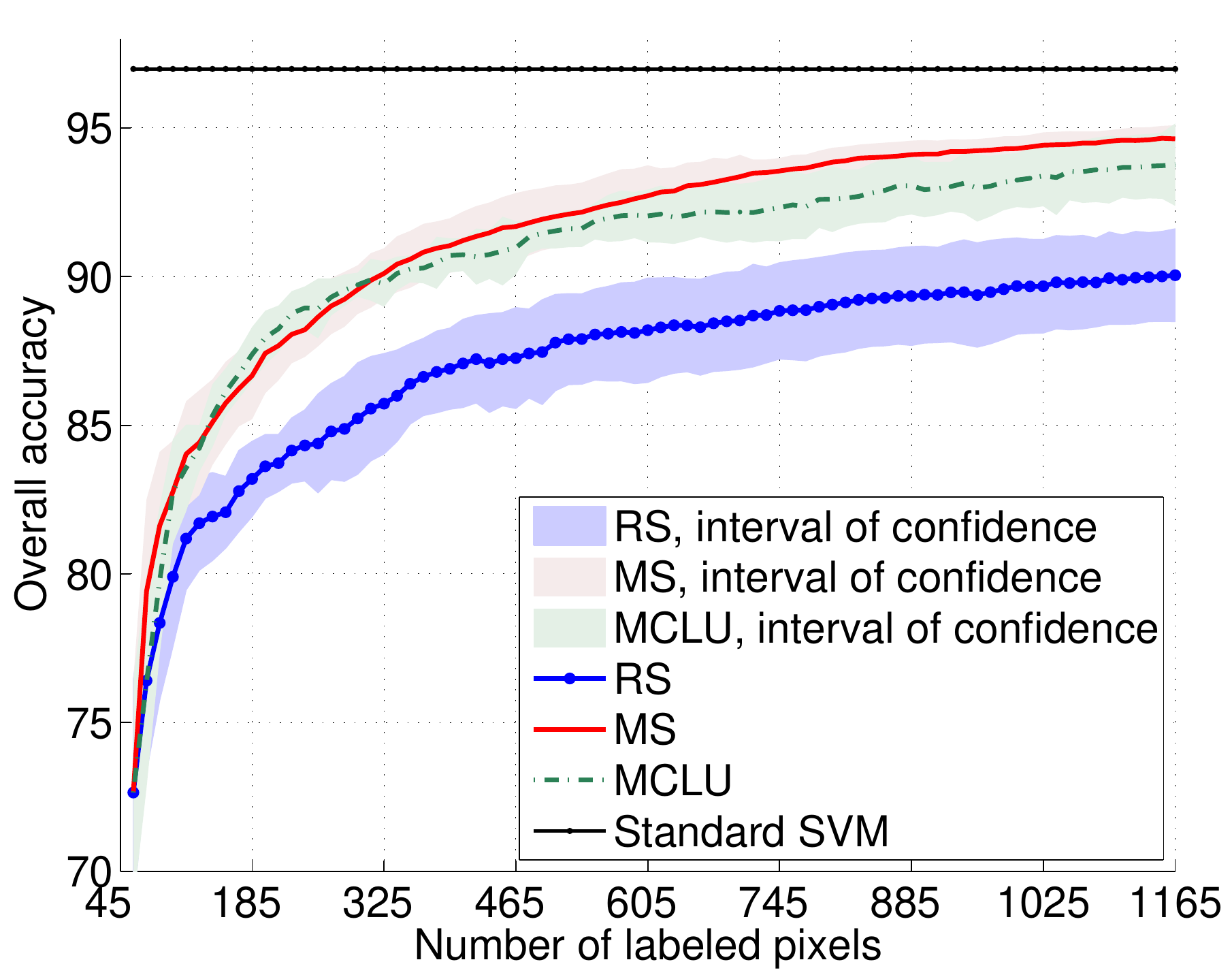}
&
\includegraphics[width=3.8cm]{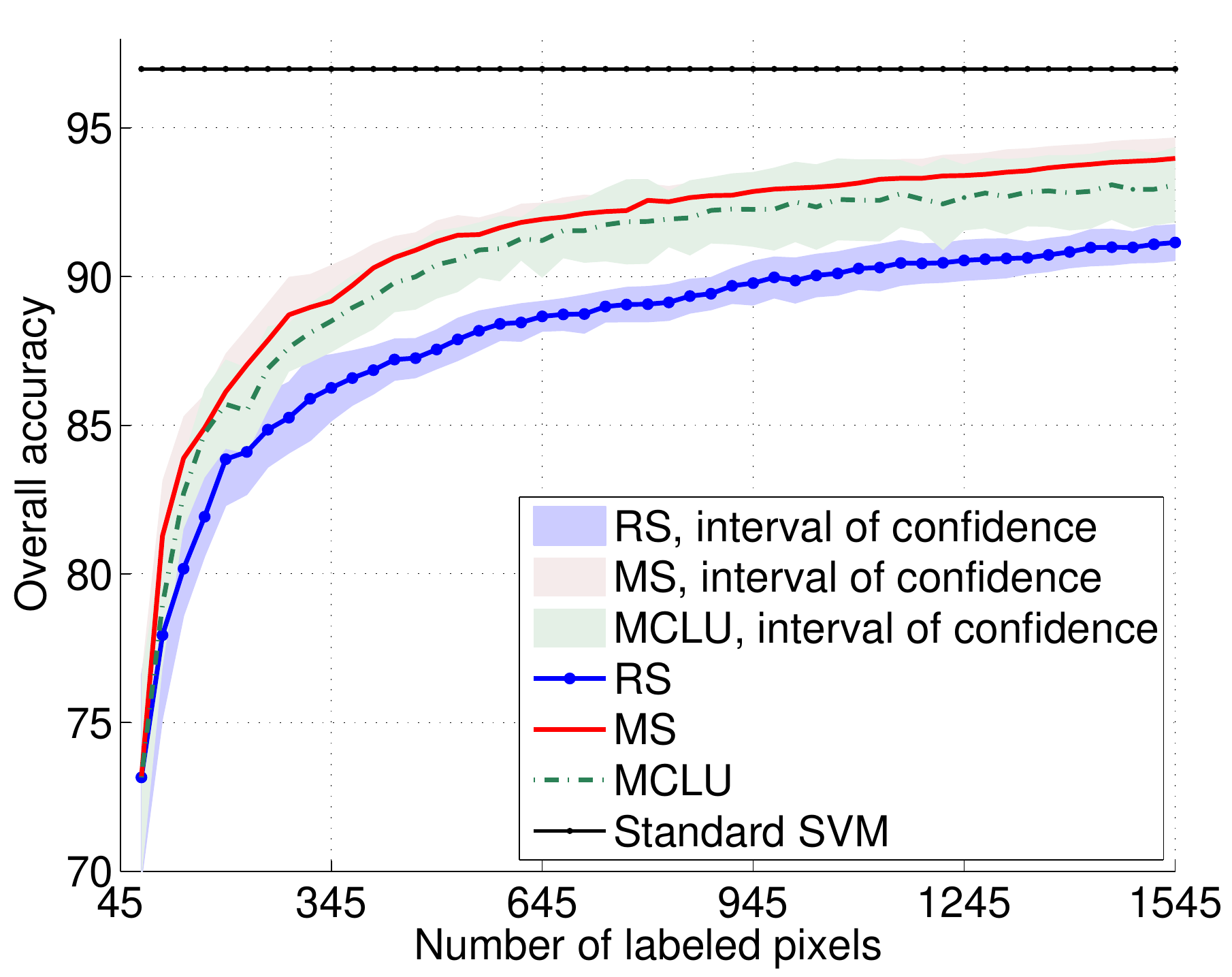} \\
&(c)&(f)\\
\end{tabular}
\caption{Large margin active learning without diversity criterion. An example comparing MS and MCLU (RS = Random Sampling).}
\label{fig:largeComp}
\end{figure}

\begin{figure}[!t]
\begin{tabular}{ccc}
& $N+5$ & $N+20$\\
\rotatebox{90}{ Pavia ROSIS}
&
\includegraphics[width=3.8cm]{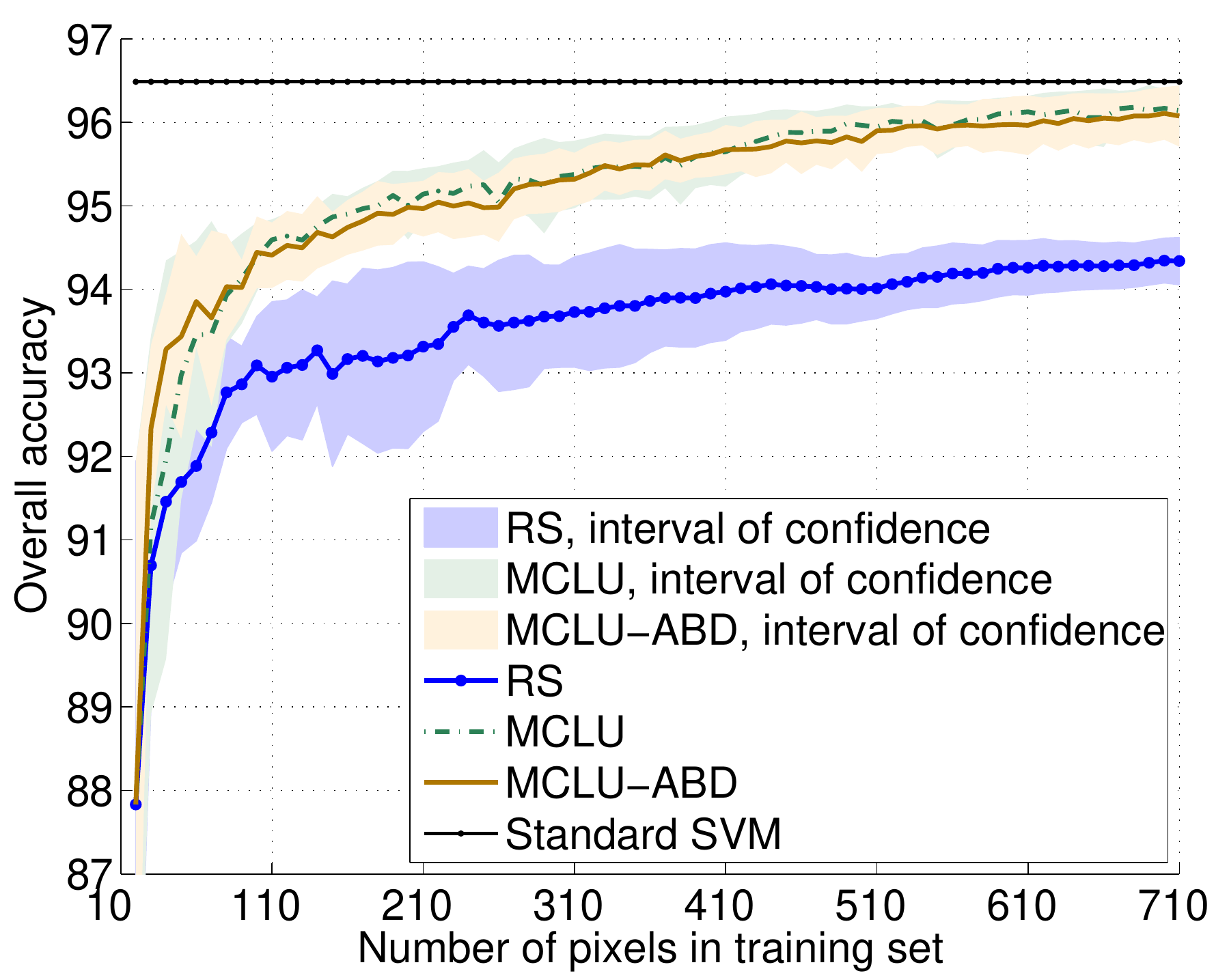} 
&
\includegraphics[width=3.8cm]{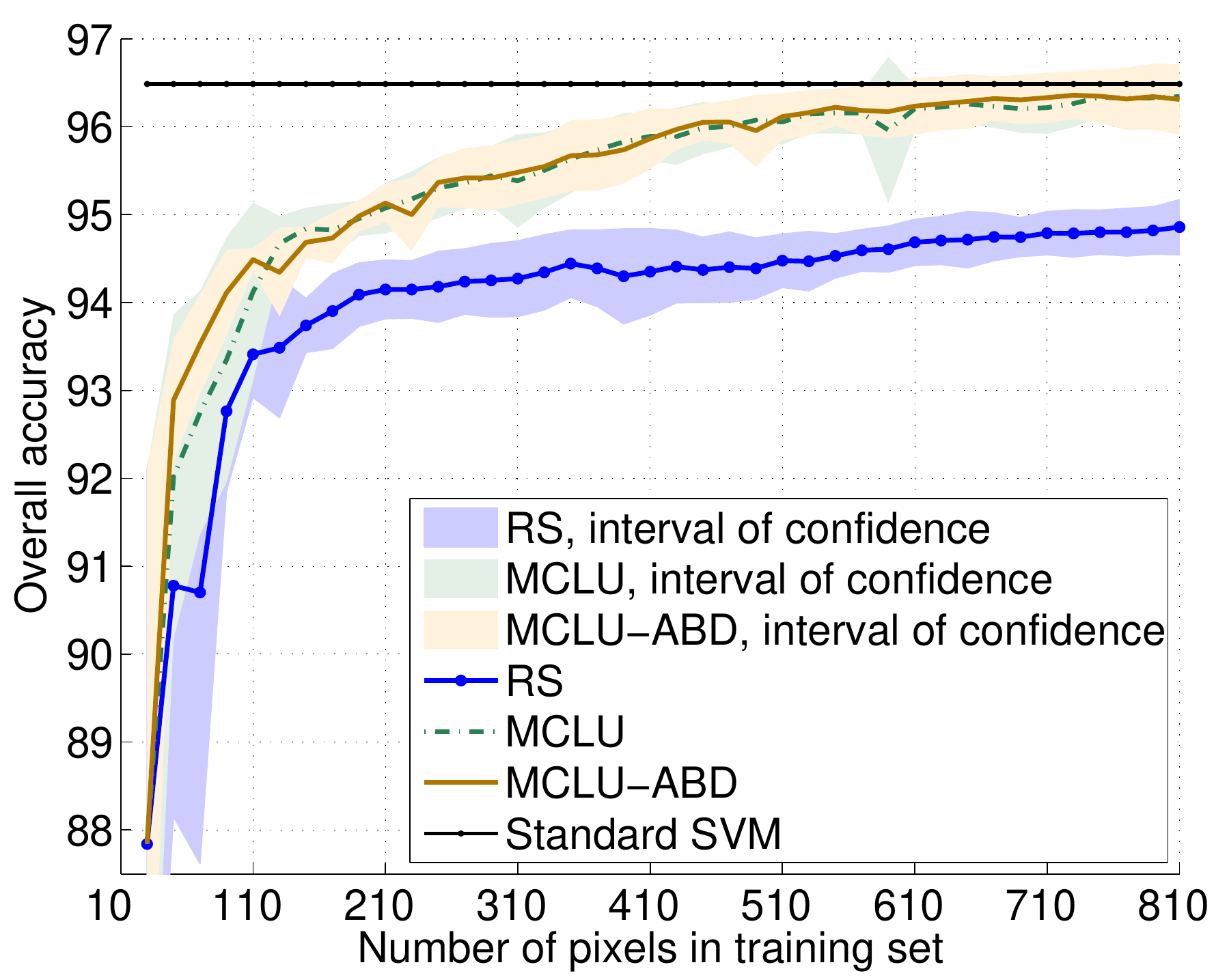} \\
&(a)&(d)\\

\rotatebox{90}{ Indian Pines AVIRIS}
&
\includegraphics[width=3.8cm]{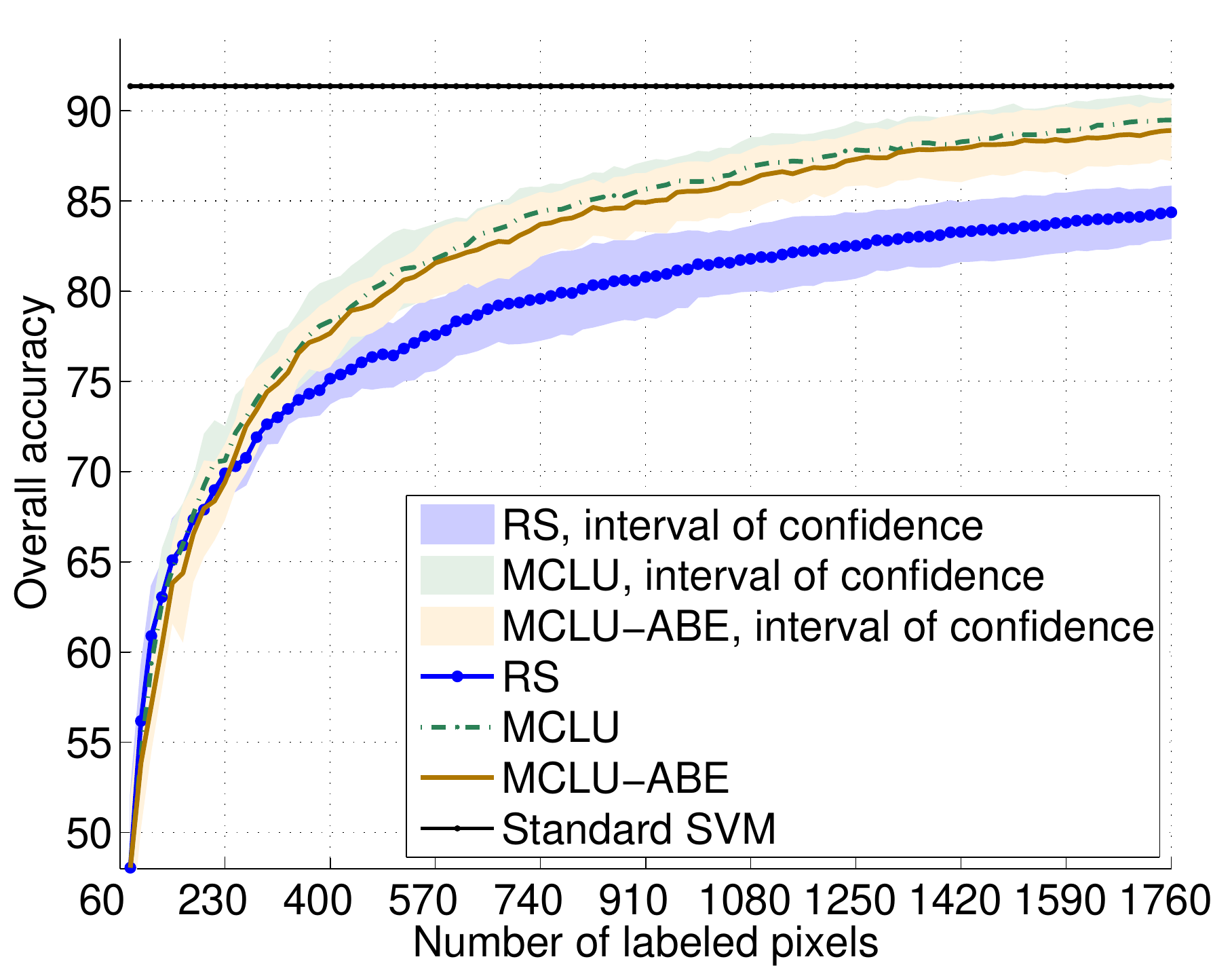} 
&
\includegraphics[width=3.8cm]{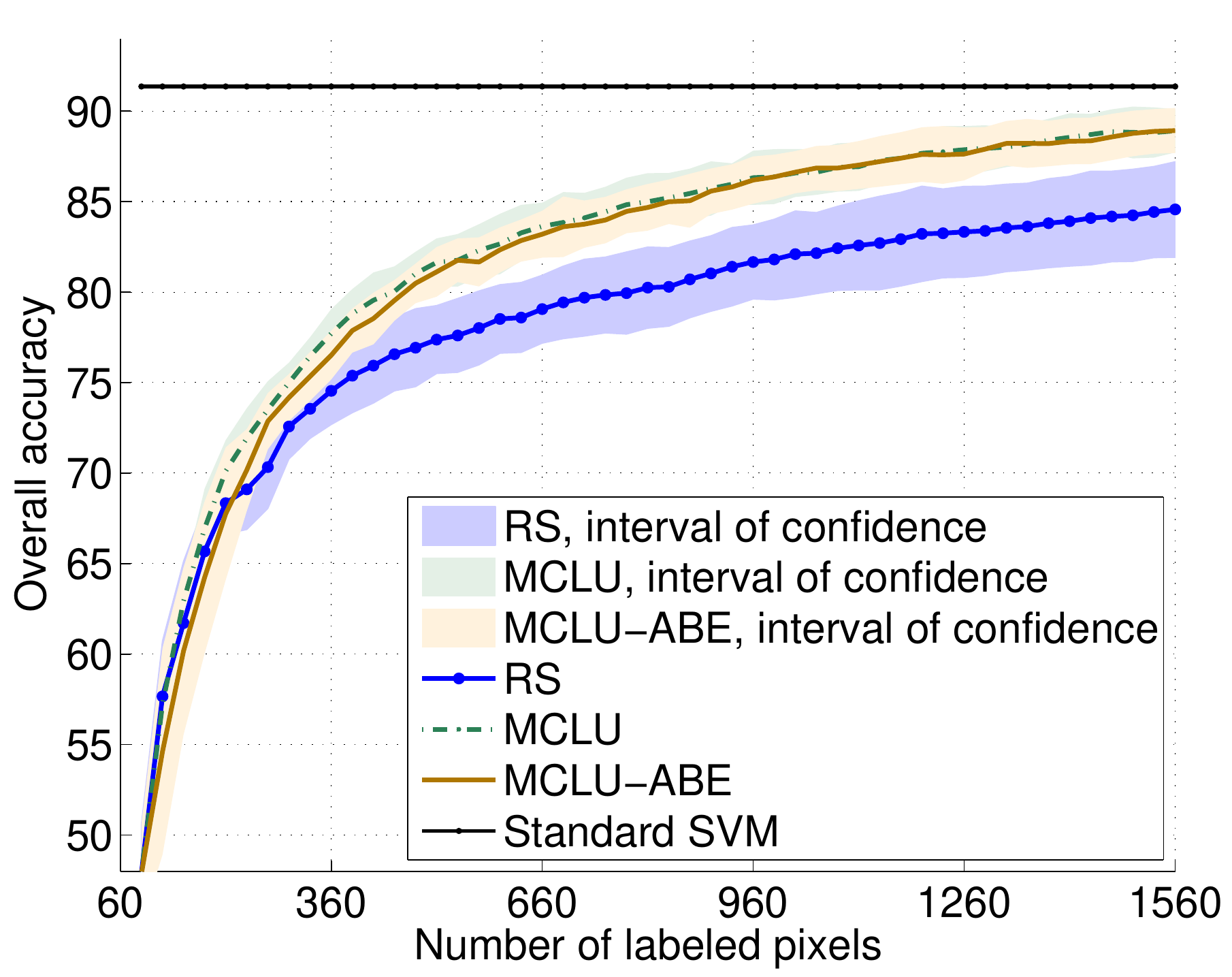}\\
&(b)&(e)\\

\rotatebox{90}{ Zurich QuickBird}
&
\includegraphics[width=3.8cm]{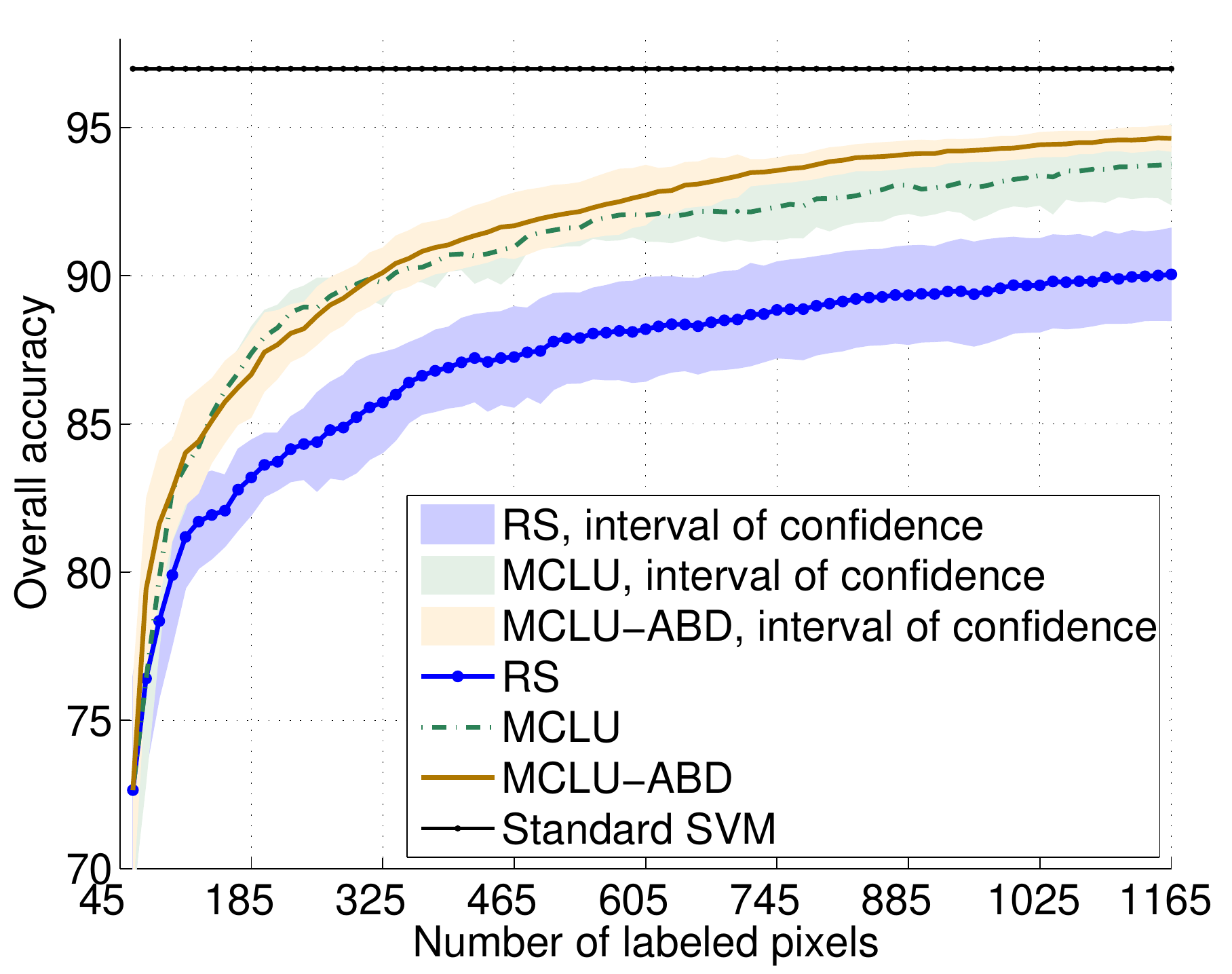}
&
\includegraphics[width=3.8cm]{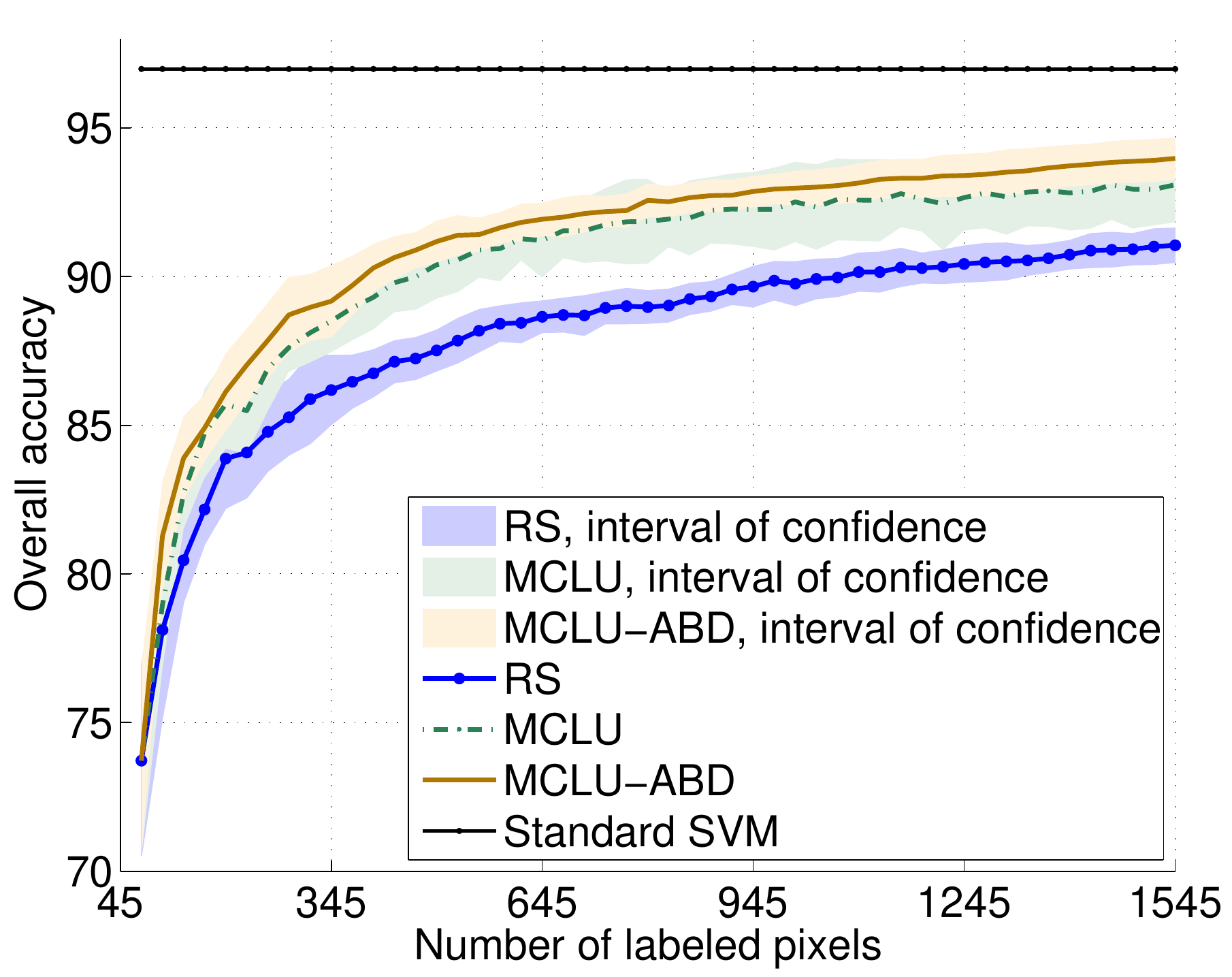} \\
&(c)&(f)\\
\end{tabular}
\caption{Effect of diversity criteria on large margin active learning. An example comparing MCLU and MCLU-ABD (RS = Random Sampling).}
\label{fig:diver}
\end{figure}

Regarding this family, Figs.~\ref{fig:largeComp} and~\ref{fig:diver} illustrate two concepts regarding the two stages of large margin heuristics: the uncertainty and diversity criteria. Figures~\ref{fig:largeComp} compares the MS and MCLU criteria and shows that both describe the uncertainty of the candidates in a similar way. Therefore, both can be used for efficient active learning. The use of a diversity criterion  seems to slightly improve the quality of the results (Fig.~\ref{fig:diver}): except for the AVIRIS image -- well known for the high degree of mixture of its spectral signatures -- a spectral diversity criterion such as MCLU-ABD efficiently increases performances with little added computational cost. None of the solutions obtained with the inclusion of the diversity criterion degrade the ones relying on the uncertainty assumption only.

\begin{figure}[!t]
\centering
\begin{tabular}{cc}
\includegraphics[width=3.8cm]{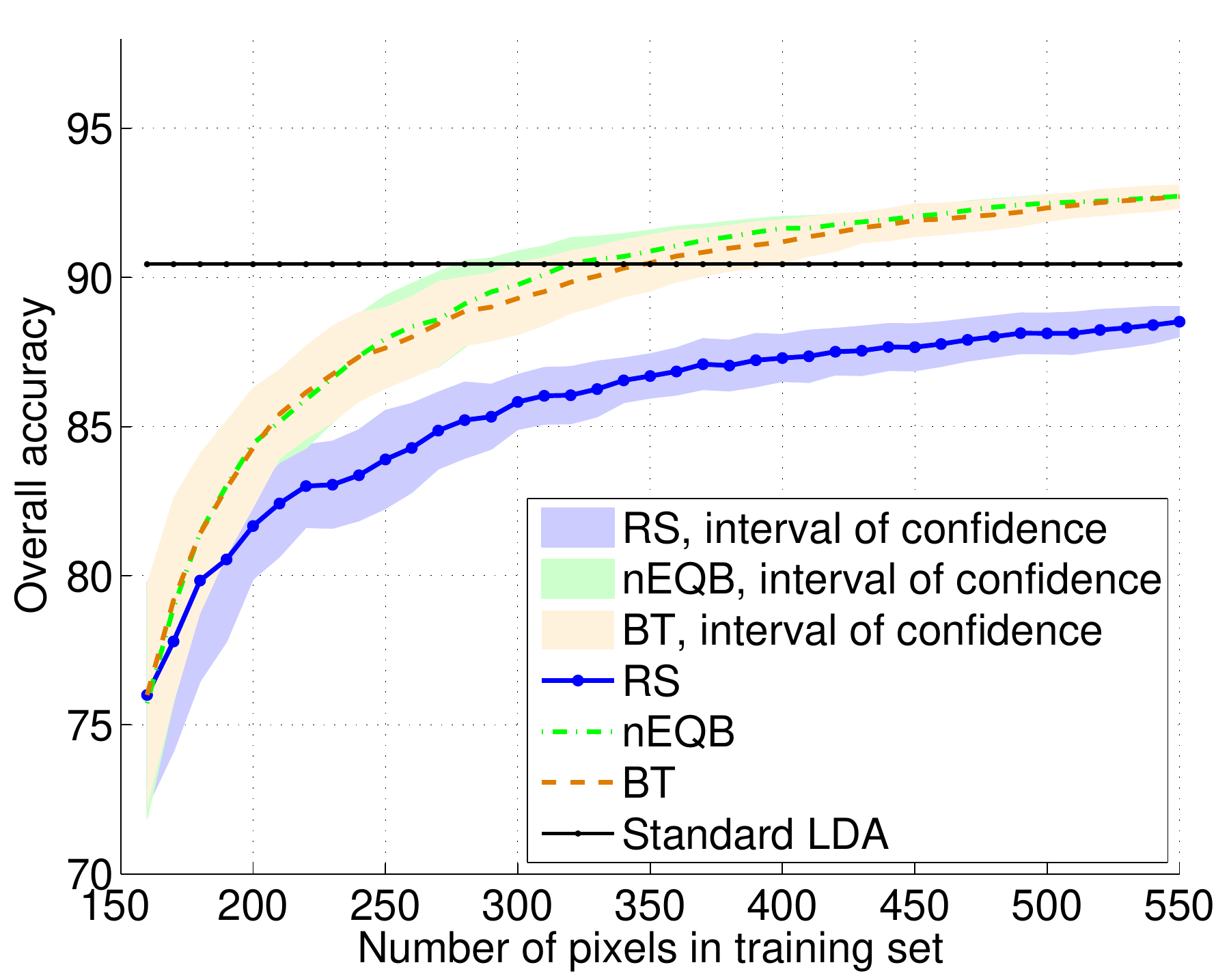} &
\includegraphics[width=3.8cm]{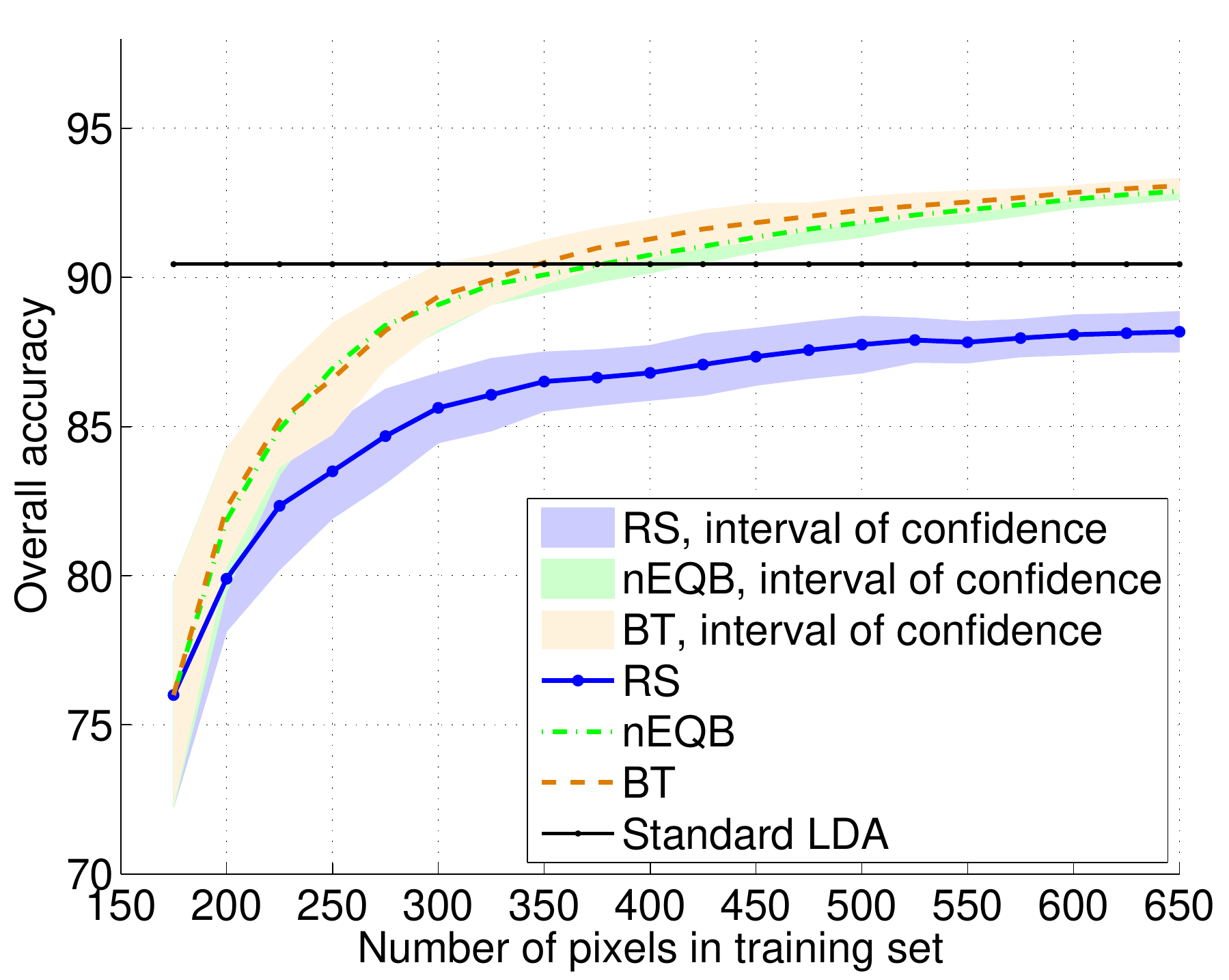} \\
(a) $N+5$ & (b) $N+20$\\
\end{tabular}
\caption{Committee-based and posterior probability heuristics trained with  LDA classifiers on the Pavia ROSIS image (RS = Random Sampling).}
\label{fig:LDA}
\end{figure}

As stated above, other heuristics must be used for other classifiers. Figure~\ref{fig:LDA} compares the $n$EQB and BT heuristics applied to the Pavia image using LDA. 

In this case, both heuristics perform similarly and show a very interesting behavior: active sampling helps the LDA to estimate the subspace that better separates data. In fact, when sampling randomly, noise and outliers can make the estimation of the Fisher's ratio biased, resulting in a suboptimal linear combination of variables in the decision function. Sampling and assigning correct labels to the pixels returned by these heuristics help estimating the correct \emph{per-class} extent (covariance) and position (mean). From 330 pixels up, the standard LDA result is improved by the active learning training sets, providing more harmonious solutions that allow a better generalization. 

Summing up, when using methods other than large margin-based algorithms, performances of the heuristics are similar and the choice must be driven by the specific constraints of time and number of iteration allowed. We will come back to these issues in the next section.

\section{Discussion}\label{sec:disc}
Throughout the experiments presented, nearly all the  algorithms compared showed fast convergence to the upper bounds represented  by the Standard SVM/LDA. At convergence, all the heuristics outperformed the random selection of pixels.  The interest of using multiclass uncertainty or adding a criterion of diversity has been demonstrated by the experiments above. Table~\ref{tab:sum} summarizes the points raised in this paper and gives a general overview of the methods reviewed. 

Large margin-based methods with diversity criterion seem the most appropriate when using SVM, while committee-based heuristic leave freedom to the user to exploit the model he is most confident with. Moreover, the user can build ensembles of classifiers exploiting the intrinsic advantages of specific classifiers for a given type of image.  Weighted committees or boosting candidates are also possible (see Section \ref{sec:committee} for some references). Probabilistic heuristics have the advantage of speed, but cannot always provide batches of samples and, if the classifier does not return  such estimates naturally, must rely on further approximations of the posterior probabilities.

However, it would not be correct to base the choice of the heuristic in a model-based fashion only. The choice of the best heuristic is problem-oriented and depends on the needs of the user in terms of time, complexity and size of the batch to be provided at each iteration. 
This section draws some guidelines to select the most appropriate heuristic. 

A first distinction could be done depending on the type of the images considered: 
\begin{itemize}
\item[-] when dealing with hyperspectral images, which are typically high dimensional, strategies taking direct advantage of the data structure should be preferred: typically, multi-view heuristics such as the AMD or the ECBD-ABD are particularly well-suited to this type of data. The first exploits cross-informations directly in the space of the spectral bands, while the second selects the samples according to spectral angles among the candidates.

\item[-] when the initial training set is very small, heuristics based on posterior probabilities should be avoided, since such estimation strongly depends on the quality of the estimation of the class statistics (typically in the case of LDA). The same holds for committees-based on bagging, especially if the bags contain a small share of the original samples.

\item[-] when dealing with complex manifolds, in which redundancy can greatly affect the quality of the sampling and of the resulting training set, approaches based on modeling of the relationships among samples in the feature space can be strongly beneficial to select pixels reflecting such complexity. The use of kernel $k$-means in the MCLU-ECBD, in hMCS or the distance to support vectors in the cSV heuristic provide solutions in this sense.

\end{itemize}

A second distinction, more related to operational requirements, is based on the type of sampling to be performed by the user~\cite{Dem10}, 

\begin{itemize}

\item[-] when working by photointerpretation (typically in VHR imaging), sampling can be done on-screen directly by the user. This allows for large amounts of iterations and can thus be solved by small batches of samples. In this case complex heuristics favoring spectral diversity are to be preferred, since the complexity of the heuristics enforcing diversity strongly increases with the size of the batch considered. 

\item[-] on the contrary, when sampling is to be done on the field (typically in hyperspectral or mid-resolution images), only a few iterations with large batches are allowed. In this case, all the heuristics seem to provide the same type of convergence and the user should prefer simple heuristics such as MCLU, BT or EQB, depending on the model used. In this case, the spatial location of samples seems to be much more important than the heuristic used: a pioneering work in this sense can be found in ~\cite{Liu08b}, where MS and BT are exploited with spatially adaptive cost and the sampling is optimized with respect to the spatial distance among the samples chosen.

\item[-] when sampling is done with moving sensors and the samples are acquired sequentially by an automatic device, batches of samples are not necessary. In this case models with small computational cost should be preferred, as they can update fast and almost instantly provide the next location to be sampled. In this case, BT and KL-max are most valuable. 
\end{itemize}

\begin{table*}[!t]
\caption{Summary of  active learning algorithms (B : binary, M : multiclass, $q$ : number of candidates, $k$ : members of the committee of learners, $S$ : batch, $SVs$ : support vectors, $\checkmark$ : yes, $\times$ : no).}
\label{tab:sum}
\begin{center}
\setlength{\tabcolsep}{1pt}
\begin{tabular}{p{1.8cm}|p{1.6cm}|c|c|c|c|c c|p{4cm}}
\hline
Family&Heuristic& Reference &Batches&Uncer-& Classifier& \multicolumn{2}{c|}{Diversity}&{Models to train}\\
&&&& tainty&&$S$&$SV$s&\\
\hline\hline

\multirow{2}{*}{Committee}&EQB&\cite{Tui09}&\checkmark&M&All&$\times$ &$\times$&$k$ models\\
&AMD&\cite{Di10b}&\checkmark&M&All&$\times$ &$\times$&$k$ models\\
\hline
\multirow{9}{*}{Large margin}&MS&\cite{Mit04}&\checkmark&B& SVM &$\times$ &$\times$&Single SVM\\
&MCLU&\cite{Dem10}&\checkmark&M&SVM&$\times$ &$\times$&Single SVM\\
&SSC&\cite{Pas10b}&\checkmark&B&SVM&$\times$ &$\times$&2 SVMs\\
&cSV&\cite{Tui09}&\checkmark&B&SVM&$\checkmark$ &$\times$&Single SVM + distances to support vectors \\
&MOA&\cite{Fer07}&\checkmark&B&SVM&\checkmark &$\times$&Single SVM + distances to already selected pixels\\
&MCLU-ABD&\cite{Dem10}&\checkmark&M&SVM&\checkmark&$\times$&Single SVM + distances to already selected pixels\\
&MCLU-ECBD&\cite{Dem10}&\checkmark&M&SVM&\checkmark &$\times$&Single SVM + nonlinear clustering of candidates\\
&hMCS-i&\cite{Vol10d}&\checkmark&M&SVM&\checkmark &\checkmark&Single SVM + nonlinear clustering of candidates and SVs\\
\hline
Posterior&KL-max&\cite{Raj08b}&$\times$&M&$$ $p(y|\x)$&$\times$&$\times$&$(q-1)$ models\\
probability&BT&\cite{Luo05}&\checkmark&M& $ p(y|\x)$&$\times$&$\times$&Single model\\
\hline

\end{tabular}
\end{center}
\end{table*}

\section{Conclusion}
\label{sec:concl}
In this paper we presented and compared several state of the art approaches to active learning for the classification of remote sensing images. A series of heuristics have been classified by their characteristics into four families. For each family, some heuristics have been detailed and then applied to {three challenging} remote sensing datasets for multispectral and hyperspectral  classification. Advantages and drawbacks of each method have been analyzed in detail and recommendations for further improvement have been worded. 
However, this review is not exhaustive and the research in the field is far from being over:  
there is a healthy and rich research community developing new heuristics for active sampling that have been or will be presented in the remote sensing and signal processing community. 

Active learning has a strong potential for remote sensing data processing. Efficient training sets are needed by the users, especially when dealing with large archives of digital images. New problems are being tackled with active learning algorithms, guaranteeing the renewal of the field. Some recent examples can be found in the active selection of unlabeled pixels for semi-supervised classification~\cite{Li10b},  spatially adaptive heuristics~\cite{Liu08b} or the use of active learning algorithms for model adaptation across domains~\cite{Jun08,Tui11a}. 
Further steps for active learning methods are the inclusion  of  contextual information in the heuristics: so far, the heuristics proposed only take advantage of spectral criteria -- or at most include contextual features in the data vector -- but few heuristics directly consider positional information and/or textures.  Another crucial issue is the robustness to noise: since they are based on the uncertainty of the pixels, current heuristics are useless for images related to high levels of noise such as SAR. This field remains, at present, totally unexplored.

\section*{Acknowledgments}
The authors would like to acknowledge prof. Paolo Gamba (Univ. Pavia) who provided the Pavia dataset, as well as the authors in \cite{Jac01} for the Indian Pines data.

\bibliographystyle{IEEEbib}
\bibliography{refsALreview}

\end{document}